\documentclass{tlp}

\pdfoutput=1

\usepackage[T1]{fontenc}

\ifx\pdfoutput\undefined
\usepackage[dvips]{graphicx}
\else
\usepackage[pdftex]{graphicx}
\fi

\newcommand{\no}{\hbox{\it not}\ }

\newtheorem{example}{Example} 

\begin{document}

\long\def\comment#1{}

\title{Reasoning Support for Risk Prediction and Prevention in Independent Living}

\author[Mileo et al.]
{A. Mileo, D. Merico and R. Bisiani \\
NOMADIS Research Lab.\\
Dept. of Informatics, Systems and Communication\\
University of Milan-Bicocca\\
viale Sarca 336/14\\
I--20126 Milan\\
E-mail: \{alessandra.mileo, davide.merico, roberto.bisiami\}@disco.unimib.it\\
}

\pagerange{\pageref{firstpage}--\pageref{lastpage}}
\setcounter{page}{1}

\maketitle

\label{firstpage}

\begin{abstract}
In recent years there has been growing interest in solutions for the delivery of clinical care for the elderly, due to the large increase in aging population.
Monitoring a patient in his home environment is necessary to ensure continuity of care in home settings, but, to be useful, this activity must not be too invasive for patients and a burden for caregivers.
We prototyped a system called SINDI (Secure and INDependent lIving), focused on i) collecting a limited amount of data about the person and the environment through Wireless Sensor Networks (WSN), and ii) inferring from these data enough information to support caregivers in understanding patients' well being and in predicting possible evolutions of their health.
Our hierarchical logic-based model of health combines data from different sources, sensor data, tests results, common-sense knowledge and patient's clinical profile at the lower level, and correlation rules between health conditions  across upper levels. The logical formalization and the reasoning process are based on Answer Set Programming. The expressive power of this logic programming paradigm makes it possible to reason about health evolution even when the available information is incomplete and potentially incoherent, while declarativity simplifies rules specification by caregivers and allows automatic encoding of knowledge. This paper describes how these issues have been targeted in the application scenario of the SINDI system.
\end{abstract}



\section{Background and Motivations}\label{sec:intro}

In the last twenty years there has been a significant increase of the average age of the population in most western countries and the number of senior citizens has been and will be constantly growing.  Living independently in their own homes is a key factor for these people in order to improve their quality-of-life and to reduce the costs for the community. For this reason there has been a strong development of computer technology for the delivery of clinical care outside of hospitals. 

For example, tele-monitoring of critical conditions (telemedicine) is becoming a common way to support medicine at-a-distance because of lower and lower costs of equipment and the savings that can be achieved. Telemedicine cannot help in prolonging Healthy Life Years (HLY) because it enters the picture when a person has already one or more critical chronic conditions. On the contrary, pervasive monitoring in the home (which has been proposed in research projects and, in a somewhat reduced form, in commercial products) has the potential of pinpointing potentially critical conditions \emph{before} they arise.  

In the last few years, many interesting systems were developed in the area of WSNs for assisted living and healthcare, among which ALARM-NET~\cite{ALARM-NET}, SAILNet~\cite{SAILNet} and CodeBlue~\cite{CodeBlue}.
ALARM-NET is a wireless sensor network designed for long-term health monitoring in assisted living and residential environments. The central design goal was to adapt the behaviour of the system, including power management and privacy policy enforcement, to the individual life patterns which are analyzed and fed into the system. The system incorporates a Circadian Activity Rhythm (CAR) analysis module used in all the reasoning about the activities performed by the users.
SAILNet proposes to apply the technology of WSNs as a non-obtrusive tool to monitor the activities of elders living in their apartments, focusing only on fall detection and pointing out that quick responses to these alarms are the critical requirement. Therefore, the project puts a lot of emphasis on the availability of WSNs. CodeBlue is a wireless communications infrastructure for critical care environments. It is designed to provide routing, naming, discovery, and security for wireless medical sensors, PDAs, PCs, and other devices that may be used to monitor and treat patients in a range of medical settings. Given our application scenario, the cited projects do not fulfill all our goals.

Rather than supporting activities~\cite{haigh02-survey} and observing behaviors~\cite{Liao04}, this paper considers a complementary view of artificial intelligence applied to home healthcare and aimed at supporting Independent Living.
If we consider the majority of commercial ``independent living'' systems, we notice that they focus on capturing medical emergencies. This can be useful but it is not the most effective way to improve the quality of life of elderly people or to lower medical costs. A more useful goal is to prevent situations that can cause drastic changes for the worse of the quality of living, e.g., falls, constant weight loss, etc. 
Pervasive and continuous monitoring of an elderly can help achieving these objectives, but it is not enough. Since, sensor-based evaluation of the health state of a person cannot be as complete as a human-based evaluation can be. We believe that reliable support for prediction and prevention is possible by integrating the deployment of pervasive sensors with expressive inference capabilities that make it possible to reason about plausible evolutions of the health state of the person while considering incomplete evaluations of aspects of the health state computed from the monitoring and their inter-dependencies.

In order to address these concerns, we have designed an independent-living support system called SINDI (Secure and INDependent lIving) that includes:
\begin{itemize}
\item A Wireless Sensor Network (WSN), composed of infrastructural nodes equipped with several sensors and a wearable monitoring device, used for gathering data about the user and the environment. 

\item A reasoning component that makes it possible to analyse the health evolution of patients in order to identify and predict what is best for the patient in his specific context~\cite{Ton01}.

\end{itemize}

\begin{figure}
\begin{center}
\includegraphics[width=0.95\textwidth,keepaspectratio]{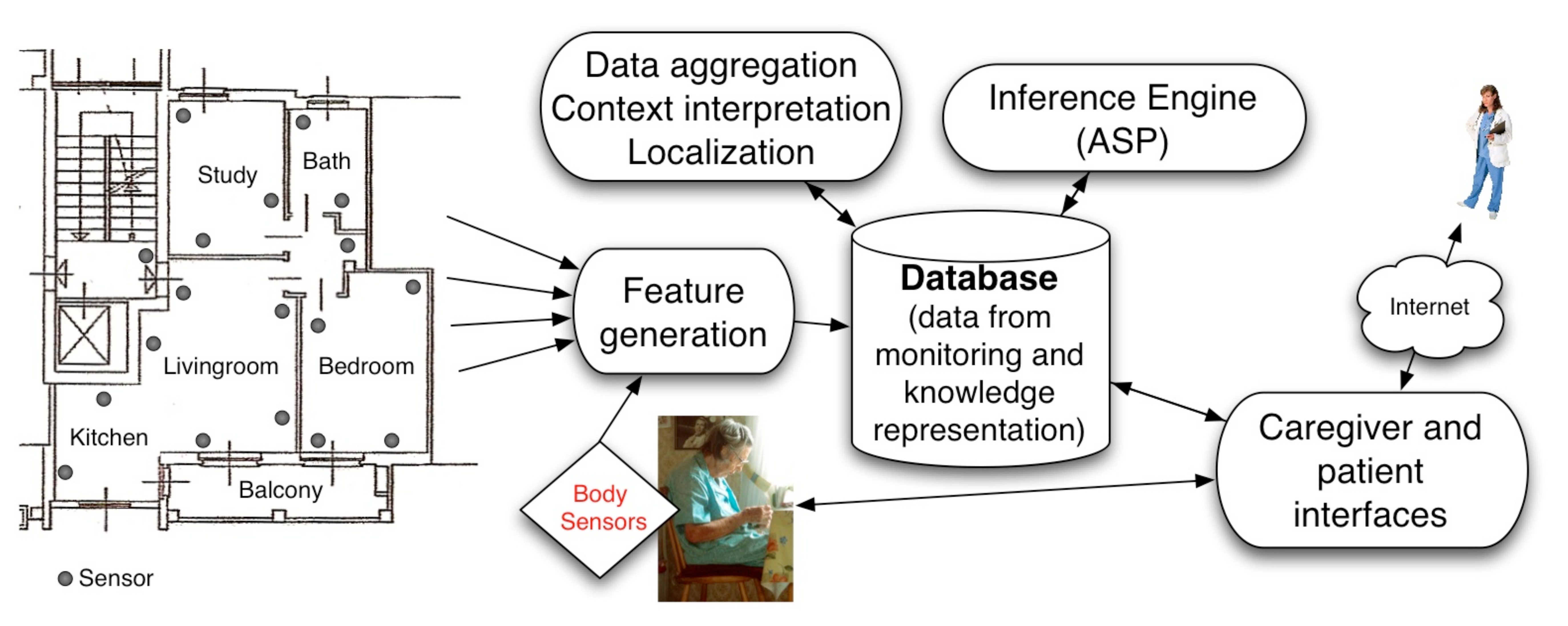}
\end{center}
\mbox{}\hrulefill
\caption{Flow of Data in the SINDI System.}
\label{fig:layers}
\end{figure}

Figure~\ref{fig:layers} shows a high-level overview of the architecture of our system and the correlation between its components. 

Wireless Sensor Networks (WSNs)~\cite{Akyildiz2002} consist of nodes that are capable of interacting with the environment by sensing and controlling physical parameters; the nodes use packet radio communication to exchange data. These networks are typically used to collect data for long periods of time without assistance. Specific scenarios for WSNs include habitat monitoring, industrial control, embedded sensing, medical data collection, building automation, fire detection, traffic monitoring, etc.

For the SINDI Wireless Sensor Network (SINDI-WSN) we designed, implemented and prototyped new WSN devices specifically conceived for people monitoring and tracking. These devices use a commercial wireless microcontroller (Jennic JN5139) with IEEE 802.15.4 radio and are equipped with heterogeneous sensors.

The SINDI-WSN is composed of:
\begin{itemize}
\item \textbf{A Master Processor}, the coordinator node of the network. It is the gateway of the network and it has storage, processing power and main memory capabilities in the ballpark of an average PC.
\item \textbf{Reference Nodes.} They are always active and connected to household power, used for network coordination but also with sensing capabilities. The presence of an always active base node in every zone simplifies data routing because the data-gathering nodes are guaranteed to always find a listening node. Therefore, they can simply send a message with the proper sensor data and quickly enter in sleep-mode without wasting precious energy.
\item \textbf{Data-Gathering Nodes.} They sense the environment data and capture specific events (e.g. opening/closing doors and windows). Moreover reference and data-gathering nodes include infrared motion and range-finder sensors used for improving the accuracy of the signal strength-based localization and tracking.
\item \textbf{A Wearable Monitoring Device.} This wearable monitoring device, called Inertial MeasUrement Device (IMUD), includes a complete six degrees of freedom Inertial Measurement Unit (IMU) and is used for user localization, movement detection and for improving the behaviour of the target tracking algorithms.
\end{itemize}

The network is organized hierarchically, as follows. 
The environment in which the user lives is divided into zones and every zone (typically a room) is controlled by one reference node. Moreover, every zone can be divided into several sensing areas where one or more environment nodes operate. The master processor manages the entire network using a middleware framework that provides several network services like: topology-control, hierarchical routing, localization, data aggregation, and communication.
This middleware has been and developed using well-known WSN algorithms~\cite{Leach00,BetterLEACH,Savvides2001} in order to carry out the data captured by SINDI-WSN and to feed the reasoning system.

We address those elderly that are clinically stable although they might be affected by chronic diseases and physical decline (more than 90\% of the population over 65 has more than one chronic disease). Since their health condition does not require constant monitoring of complex biomedical parameters, these patients do not need, and are less tolerant of, invasive sensors.

Computing the cost of the system if it were deployed as a product critically depends on the commercial model: should it be cast as a service with the hardware freely given as part of the service agreement or should it be an end-user product? Both these possibilities and many in between are realistic. Such computations do not belong in this paper but we can state that the hardware replication cost is small enough to make commercialization viable.

We use monitoring to support health evaluation, prediction, explanation, emergency detection and prevention in the same framework and provide a global representation and reasoning model for general health assessment, combining medical knowledge, patient's clinical profile, context and patient evaluation through sensor data, nonmonotonic reasoning and qualitative optimization.

The need of making the system user-centered and medically sound led us to include domain-specific medical knowledge in the reasoning phase. In this way it is possible to trace general habits and their correlation with the patient's well-being according to the evaluation methods of clinical practice.

Medical soundness and context-awareness improve the reliability of the system because the combination of different sources of information (sensors, medical knowledge, clinical profile, user defined constraints) that change over time makes the system more reliable (i.e., much better able to disambiguate situations, thus reducing false positives) and adaptive (e.g., suitable for the introduction of new available information).  

To devise an appropriate knowledge representation and reasoning model for SINDI, we considered the main classes of tasks we want our independent-living system to perform, namely:

\begin{itemize}
\item {\bf Interpretation of Context} based on a logic-based context model and consisting in i) {\it Localization} of the person in the environment and ii) {\it Identification} of simple activities (opening doors/windows, sleeping, walking, etc.).
\item {\bf Health Assessment} based on a logic model of health and consisting in i) {\it Evaluation} of significant aspects of the patient's quality of life (referred to as \emph{indicators}) including quality of the environment, quality of movement, quality of sleep, weight, use of lights, etc., and iii) {\it Evaluation} of a well-defined set of health-related factors (referred to as \textit{items}); the evaluation of each item is obtained according to the values of the indicators related to that item and the influences between each indicator and the item (see Figures~\ref{fig:model1} and~\ref{fig:model2} )\footnote{As an example, the item \emph{insomnia} is influenced by indicators representing the quality of sleep in the early night (\emph{earlyNight}), in the heat of the night (\emph{middleNight}) and in the early morning (\emph{lateNight}); the negative dependency connecting these indicators with \emph{insomnia} determines that a worsening of the indicators results in a worsening of the functional disability represented by \emph{insomnia}.}.
\item {\bf Health Evolution} based on the analysis of dependencies among items and consisting in i) {\it Prediction} of possibly-risky situations and identification of dependencies representing plausible causes, ii) {\it Explanation} of identified risks through common causalities and iii) {\it Reaction} of the system in form of predefined warnings/suggestions/actions aimed at prevention.
\end{itemize}


According to these views, there are three components of the knowledge model and three aspects in which reasoning is involved: the first one is related to the use of sensor data and common-sense reasoning for context interpretation, the second one refers to the (partial) evaluation of indicators according to the results of context interpretation and the (partial) evaluation of items according to related indicators, while the last one is related to understanding and predicting the evolution of the health state according to the dependency graph, in order to provide appropriate feedback and support clinicians' understanding.

Section~\ref{sec:asp} reviews the logical framework for knowledge representation and reasoning used in SINDI.
Section~\ref{sec:ctx-int} describes the knowledge representation model of SINDI with respect to the context and the elderly care, while Section~\ref{sec:reasoning} gives details about SINDI's reasoning tasks in terms of context interpretation, context-dependent evaluation and health assessment. Section~\ref{sec:eval} contains a preliminary evaluation.

\section{The Logical Framework}\label{sec:asp}


The declarative logical framework we use for Knowledge Representation and Reasoning in SINDI is that of Answer Set Programming (ASP), based on the \emph{stable model} semantics for Logic Programs proposed by Gelfond and Lifschitz~\cite{GelLif88,Marek89,Nie99}. 

Compared to pure statistical approaches, logic inference based on ASP is highly expressive and computationally more performant because it can deal with first-order representations, which are much richer than the propositional ones characterizing probabilistic inference. Furthermore, ASP can deal with incomplete information and common-sense reasoning using defaults. Cardinality and weight constraints together with optimization techniques are also interesting features for our application, in that they can be used to model different degrees of uncertainty~\cite{BreNieSyr02,Nie02,Dlv06,Geb07b}: given the incompleteness of available knowledge, we may need to use both optimization criteria to select the best candidate solutions according to both qualitative and quantitative criteria.
Declarativity also represents a desirable feature because it allows the automatic encoding of medical knowledge, thus making the system easily extensible and medically sound.

Before we describe our Knowledge Representation Model and Reasoning Algorithms, we want to recall some basic ASP definitions. In ASP a given problem is represented by a logic program whose results are given in terms of \emph{answer sets}. 


A logic program $P$ is a finite set of rules $r_i$ of the form
\begin{equation}\label{eq:rule}
r_i:\ L_0\ \leftarrow\ L_1,\ldots,\ L_m,\ \no\ L_{m+1},\ldots,\ \no\ L_n\ .
\end{equation}

where $L_i\ (i=0..n)$ are literals, $\no$ is a  logical connective called
{\em negation as failure} and $n \geq m \geq 0$. 
We define $L_0=head(r)$ as the \emph{head} of rule $r_i$, and $body(r_i)=L_1,\ldots,\ L_m,\ \no\ L_{m+1},\ldots,\ \no\ L_n$ as the \emph{body} of $r_i$. Furthermore, let $body^+(r_i)=\{L_1,\ldots,L_m\}$ and $body^-(r_i)=\{\no\ L_{m+1},\ldots,\ \no\ L_n\}$.
Rules $r_i$ with $head(r_i)=\emptyset$ are called \emph{integrity constraints}, while if $body(r_i) = \emptyset$, we refer to $r_i$ as {\em a fact}.

An interpretation is represented by the set of atoms that are true in it. 
A \emph{model} of a program $P$ is an interpretation in which all rules of $P$ are true according to the standard definition of
truth in propositional logic.
Apart from letting '$,$' stand for conjunction, this implies treating rules and default negation by~$\mathit{not}$ as implications and classical negation, respectively.
Note that the (empty) head of an integrity constraint is false w.r.t.\ every interpretation, while the empty body is true w.r.t.\ every interpretation.
Answer sets of~$P$ are particular models of~$P$ satisfying an additional stability criterion.
Roughly, a set~$X$ of atoms is an answer set, if for every rule of form~(\ref{eq:rule}), $L_0 \in X$ whenever $L_1,\dots,L_m$ belong  to~$X$ and no $L_{m+1},\dots,L_n$ belongs to~$X$.

Formally, an \emph{answer set}~$X$ of a program~$P$ is a minimal (in the sense of set-inclusion) model of
\[
\{
head(r_i) \leftarrow body^+(r_i)
\mid 
r_i \in P,body^-(r_i)\cap X=\emptyset
\}.
\]

Although answer sets are usually defined on ground (i.e., variable-free) programs, the rich modeling language of ASP allows for non-ground problem encodings, where rules with variables (upper case names) are taken as a shorthand for the sets of all their ground instantiations.
Grounders, such as \emph{gringo}~\footnote{\emph{A user's guide to gringo, clasp, clingo, and iclingo}. http://potassco. 
sourceforge.net} and \emph{lparse}~\footnote{Lparse 1.0. http://www.tcs.hut.fi/ 
Software/smodels/lparse.ps.gz}, are capable of combining a problem encoding and a problem instance (typically a set of ground facts) into an equivalent ground program, which can then be then processed by one of the available ASP solvers.
In our implementation of SINDI, answer set programs are grounded using Gringo~\cite{Geb07b} and interpreted using the Clasp~\cite{Geb07a} solver.

\section{Knowledge Representation Model for the Home Healthcare Domain}\label{sec:ctx-int}

Given that a well designed model is crucial to make effective reasoning possible, we carefully formalized SINDI's knowledge by using domain experts, published data and common-sense information.

The home healthcare domain is characterized by two main aspects: the context and the health assessment.
These two aspects are strictly related in the home monitoring scenario, because the capability of identifying meaningful information about the context in which a person lives is a critical issue for health assessment via monitoring.

In this section we want to separately describe the knowledge representation model of SINDI in terms of these two domain aspects.

\subsection{Model of Context in Home Environments}

\begin{table}
\caption{Generic Spatial Relations Among Entities.}\label{tab:spatial}
\begin{tabular*}{\textwidth}{lllp{10.5cm}}
\hline
\textbf{Relation Name} & \textbf{Object} & \textbf{Reference Object} & \textbf{Relation Type}\\
\hline
\textit{personIn/personOut} & Person & \{Room,Area\} & generic directional relation\\
\textit{in/out} & Area & Room & generic directional relation\\
& Object & \{Room, Area\}& \\
\textit{near/far} & Person & Object & generic distance relation\\
\textit{connected} &  \{Room,Area\} & \{Room,Area\} & generic directional relation\\
\hline 
\end{tabular*}
\end{table}

While most of the implemented context-representation models are domain-dependent and do not support powerful inference, our declarative logic-based description of the domain aims at providing a representation of context-dependent data that is both general and with good computational properties.
Other interesting properties of our logic-based solution are:
\begin{itemize} 
  \item {\it readability} and simplicity of the problem specification,
  \item {\it flexibility} with respect to the sources of knowledge (heterogeneous sensors can be taken into account),
  \item {\it modularity} in the specification of the problem that describes properties of a desired solution, and
  \item {\it expressivity} of the modeling language and computational efficiency of the inference engines. 
\end{itemize}

To fulfill these requirements, we use a high level description of home environments in terms of rooms, areas, objects, properties, relations and observations. The space is represented as a grid where each cell is identified by coordinates $X,Y$ and has some properties (being in a room, being a wall, being a passage, etc.). The resulting context specification is then mapped into a set of logic predicates in the Answer Set Programming (ASP) framework (see Section~\ref{sec:asp} for details). 
In addition to the description of the model, a limited set of consistency constraints is introduced to make sure that observations and context interpretation are coherent. As an example, in the \emph{localization} task, the person cannot be in a cell that has the property of being a wall. These constraints are also mapped into ASP.

Properties used in the model take into account generic spatial properties rather than describing geometric spatial relations between objects. This results in greater generality because we do not need a complete physical description of the environment.
In addition, while data gathered by the sensors are processed and aggregated according to specific algorithms for feature analysis, the information available at upper levels is filtered by the abstraction. This enables us to represent meaningful information as properties of objects, rooms or areas, keeping the model independent of  sensors' characteristics and positioning.


Our modelling approach is similar to what we would obtain by using an ontology, with the difference that the ASP reasoning enhances the expressivity and computational efficiency of the model. 
We are aware of the fact that research efforts are converging toward the combination of nonmonotonic reasoning and ontology-based knowledge representation, but available implementations are still domain dependent and formal issues need to be further explored.
For this reason we decided to encode our contextual information directly into logic predicates, that can be easily mapped into an (existing or new) ontology if needed.

Previous investigation of context models has indicated that there are certain entities in a context that, in practice, are more important than others. These are location, identity, activity and time~\cite{Ryan98,Schilit94}.
In fact, in the context of home monitoring, the more intuitively relevant aspects of a context are: \textit{where} you are, \textit{who} you are (clinical profile), \textit{which} resources you are using, \textit{what} you are doing and \textit{when}. 

In order to represent this information in our model we identified four main types of entity: \textit{Person}, \textit{Room}, \textit{Area} and \textit{Object}.

\begin{figure}[t]
\includegraphics[width=0.85\textwidth,keepaspectratio]{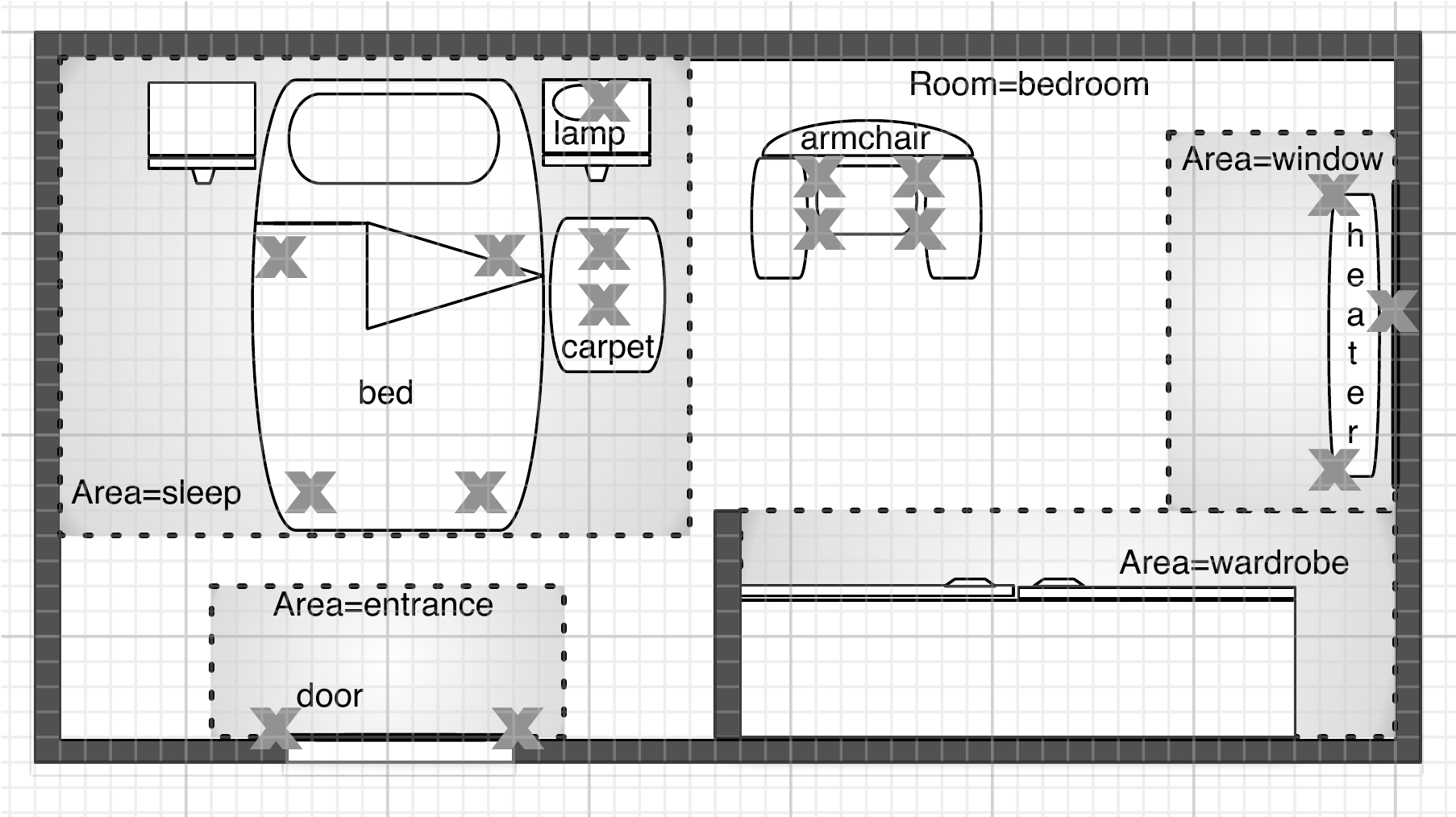}
\caption{Example: modelling a bedroom}\label{fig:bedroom}
\end{figure}

We also define a small subset of generic spatial relations among entities, summarized in Table~\ref{tab:spatial}%
\footnote{Note that the spatial inclusion of areas $A_1,\ldots,A_n$ in a room $R$ is such that $\bigcup_{i=1}^n A_i\ \subset\ R$.}.
An example of how a bedroom can be represented in our context model is illustrated in Figure~\ref{fig:bedroom}.

Attribute values may come from i) external knowledge (observed values for attributes of the Person entity), ii) opportunely aggregated sensor data (all other attributes) or iii) results of the inference process (inferred values for attributes of the Person entity, when observed values are not available).

Most of the values for attributes associated to rooms, areas and objects are the result of a process that maps numerical sensor data into meaningful thresholds. The thresholds are derived both from objective considerations, e.g. a given temperature might be too hot for the human body to survive, and from patient-profile and environment related considerations, e.g. a southern Italian and a British person might have a very different idea of what is comfortably hot or cold.
 
Values of both attributes and spatial relations are dynamic and need to be associated to an interval of time or to a discrete time point. In this way, the reasoning system can take into account their evolution during context interpretation in order to understand what the person is doing (in terms of movements) and where the person is (localization).

Lack of specificity w.r.t. the spatial relations is compensated by the inference process: reasoning about relations and attribute values may help inferring new information or defeating previous conclusions based on data fusion algorithms.

Each type of entity is associated to a specific set of attributes as follows:
\begin{itemize} 
  \item attributes associated to the \emph{Person} entity concern the evaluation of functional disabilities, dependency in performing Activities of Daily Living (ADL), assessment of risks, results of specific tests, medications, diseases, weight and attributes related to movement including motion, posture and direction of motion (see Table~\ref{tab:profile});
  \item attributes associated to the \emph{Room} and \emph{Area} entities include temperature, light, humidity, sound, presence of a moving entity, presence of gas/smoke (see Table~\ref{tab:RA});
  \item attributes associated to the \emph{Object} entity include light, temperature, humidity and sound of the object, state of the object and any other property that the object can detect using specific sensors such as waterflows (see Table~\ref{tab:Object}).
\end{itemize}

\begin{table}
\caption{Attributes representing information about the Person entity.}\label{tab:profile}
\begin{tabular*}{\textwidth}{llp{10.5cm}}
\hline
\textbf{Signature} & \textbf{Domain Values} & \textbf{Description} \\
\hline
<Func,Val> & Func=\{gait, balance, vision, cognition, sleep, & Functional\\
& nutrition \} & disability\\
& Val =\{absent, mild, moderate, severe\} & \\
\hline
<Adl,Val> & Adl=\{mobility, dress, eat\} & ADL\\
& Val=\{ok, needy, dependent\} & dependency\\
\hline
<Risk, Val> & Risk=\{fall, depression, isolation, frailty\} & Risk \\
& Val =\{absent, mild, moderate, severe\} & assessment\\
\hline
<Test, Val> & Test=\{amtest, minimental, nutritionTest, & Test\\
& clockDrawing, gds, audiometric, visual\} & results\\
& Val=\{ok, mild, moderate, severe\} & \\
\hline
<Drug, Val> & Drug=\{benzodiazepine, antidepressant, diuretic, & Medications\\
& antiarrythmic, anticonvulsant, neuroleptic, & \\
& ssri, anticholinergic\} & \\
& Val =\{yes, no\} & \\
\hline
<Disease, Val> & Disease=\{visual\_impairment, vertigo, artrosis, & Diseases\\
& feet\_disorders, edentia, arrythmia, arthritis, & \\
& hemiparesis, parkinson, acute\_pathologies, & \\
& alcoholism, hypotension, ipoacusia, ipovisus & \\
& distyroidism, malnutrition, affective\_disorders, &\\
& cognitive\_impairment\} & \\
& Val=\{ok, mild, modearte, severe\} & \\
\hline
<wgt, Val> & Val=\{1..300\} & Weight (kg)\\
<height, Val> & Val=\{100..250\} & Height (cm)\\
<gender, Val> & Val=\{m,f\} & Gender\\
\hline
<motion,Val,P$^1$> & Val=\{walk, still, null$^{2}$\}, P=\{0..100\} & Motion activity\\
<posture,Val,P$^1$> & Val=\{sit, lay, stand, null$^{2}$\}, P=\{0..100\} & Posture of the person\\
<dir,Val,P$^1$> & Val=\{turn, straight, null$^{2}$\}, P=\{0..100\} & Direction of motion\\
\hline
\end{tabular*}
\begin{tabular}{ll}
$^1$ & \footnotesize{Value ``null'' is related to the fact that no signal is received from sensors detecting movement.}\\
$^2$ & \footnotesize{Parameter ``P'' represents data reliability, and it is computed by the algorithms used for}\\
 & \footnotesize{feature extraction.}
\end{tabular}
\end{table}

\begin{table}
\caption{Attributes representing information about the Room and Area entities.}\label{tab:RA}
\begin{tabular*}{\textwidth}{llp{10.5cm}}
\hline
\textbf{Attribute Name} & \textbf{Domain Values} & \textbf{Description} \\
\hline
ambientLight & \{dark, shadow, clear, bright\} & brightness of the environment\\
ambientLightType & \{natural, artificial\} & nature of the light\\
ambientHumidity & \{dry, medium, wet, superWet\} & humidity level\\
ambientTemperature & \{cold, chilly, warm, hot, burning\} & temperature\\			   
ambientSound & \{mute, mild, medium, noisy\} & noise level\\
presence & \{yes, no\} & presence of a moving entity\\
noxiousGas & \{yes, no\} & presence of noxious gas\\
smoke & \{yes, no\} & presence of smoke\\
\hline
\end{tabular*}
\end{table}

\begin{table}
\caption{Attributes representing information about the Object entity.}\label{tab:Object}
\begin{tabular*}{\textwidth}{llp{10.5cm}}
\hline
\textbf{Signature} & \textbf{Domain Values} & \textbf{Description} \\
\hline
objectLight & \{dark, shadow, clear, bright\} & light produced by the object\\
objectLightType & \{natural, artificial\} & nature of the light\\
objectTemperature & \{hot, cold\} & meaning depends on object \\
objectSound & \{noSound, regularSound, loudSound\} & meaning depends on object \\
switch & \{open, closed\} & state of doors/windows objects \\
state & \{on, off\} & state of on/off devices\\
filteredLoad & \{0..300\} & weight measurement from\\
& & mat sensors (load-cells)\\ 
loadVolatility & \{stable, mildlyUnstable, veryUnstable\} & volatility of filteredLoad\\
waterflow & \{yes, no\} & water flowing through the object \\
\hline
\end{tabular*}

\end{table}

The expressive power of ASP is used in the context interpretation phase in order to disambiguate unclear situations as much as possible by using defaults, nondeterministic choice and constraints over the solutions (see Section~\ref{sub:ctx-reasoning} for details).
As an example, if you consider the problem of tracking the person on the grid, the reasoning system takes as input the result of a traditional particle filter based only on radio signal strength measurements. These results are combined with a model of movement and opportunely aggregated data from infrared sensors and range finders%
\footnote{Values of infrared sensors and range finders are mapped into facts of the form $sense(S,X,Y,T)$, where S=\{motion, distance\}; the presence of such facts in the problem instance indicates that the correspondent sensor data has been detected in a give cell $X,Y$ at time $T$}. 
Multiple preference criteria are applied in order to select the best solution according to the more reliable results of the particle filter, the best move and, finally, the most coherent position with respect to all available sensor data.
Non deterministic choice is used to generate possible solutions, while the combination of optimization criteria are used to find the best candidates (see Section~\ref{sub:ctx-reasoning}). 

When the results of the particle filter are not available, tracking the person on the grid becomes more difficult and the space size of the solution can be huge. This happens in particular when the output of the particle filter is missing for several sequential time stamps, since the model of movement produces a high number of possibilities. In this case, values obtained by range finders and infrared sensors can be combined with information about the state of lights or the state of objects to obtain a less fine grained localization in a room/area rather than in a cell $X,Y$.


\subsection{Model of Health in Elderly Care}\label{sub:model}

Health-related aspects of elderly care are modelled in SINDI by a two-layered graph: the bottom layer includes what we call \emph{indicators} and the upper layer includes what we call \emph{items}. 

The relation between items and indicators is the following: each indicator can contribute to the evaluation of one or more items when no direct evaluation of the item itself is available; while indicators do not have any mutual dependency or correlation among them, items are correlated by dependency relations indicating how a change in the value of an item may impact values of other items and how.
Details about how this graph-like structure is used in the reasoning process will be given in Section~\ref{sec:reasoning}.

Indicators and items that are meaningful have been identified according to the medical practice in health assessment of elderly \cite{elderly95} and encoded in our declarative framework as logic facts. A reduced list of indicators evaluated by the system is provided in Table~\ref{tab:ind} while items are grouped according to the class they belong to, as described later in this section.


As one can easily understand, evaluating items is not always trivial: the incompleteness and heterogeneous nature of collected data and the need for state-based context interpretation in dynamic systems suggest that nonmonotonic reasoning techniques can be a powerful tool for effective context-dependent reasoning under the assumption of incomplete knowledge.

\begin{table}
\caption{Indicators of Well-Being according to their differential evaluation.}\label{tab:ind}
\begin{tabular*}{\textwidth}{llp{10.5cm}}
\hline
\textbf{Indicator Name} & \textbf{Source} & \textbf{Description} \\
\hline
\textit{tests} & human input & Test results\\
\textit{drugs} & human input & Drug intake\\
\textit{diseases} & human input & Level of present diseases\\
\textit{diet} & human input & Diet type\\
\textit{sitting} & aggregation & Quality of movements\\
\textit{standing} & aggregation & Quality of movements\\
\textit{laying} & aggregation & Quality of movements\\
\textit{turning} & aggregation & Quality of movements\\
\textit{walking} & aggregation & Quality of movements\\
\textit{numSteps} & aggregation & Number of steps\\
\textit{walkSpeed} & aggregation & Walking speed\\
\textit{walkTime} & aggregation & Percentage of walking time\\
\textit{sitEquilibrium} & aggregation & Quality of movements\\
\textit{standEquilibrium} & aggregation & Quality of movements\\
\textit{wgt} & aggregation & Weight expressed in kilograms\\
\textit{lightUsage} & inference & Evaluate correct usage of lights\\
\textit{earlyNight} & inference & Quality of sleep (11:00 p.m. to 11:59 p.m.)\\
\textit{middleNight} & inference & Quality of sleep (12:00 to 2:39 a.m.) \\
\textit{lateNight} & inference & Quality of sleep (3:00 to 4:59 a.m.)\\
\textit{dayActivity} & inference & Level of activity vs. inactivity period during the day\\
\textit{tempQual} & inference & Quality of the environment w.r.t. temperature\\
\textit{humQual} & inference & Quality of the environment w.r.t. humidity\\
\textit{brightQual} & inference & Quality of the environment w.r.t. light\\
\textit{socialAct} & inference & Quantity of social interactions\\
\hline 
\end{tabular*}
\end{table}



A careful analysis of the elderly care in home settings suggests that health-related items can be classified into four classes: State representing the well-being of the person and the environment in which she lives, Functionalities representing functional disabilities of the person monitored, Activities of Daily Living (ADLs) representing her dependence in performing daily activities, and Risk Assessment characterizing risky conditions. 

In the remaining part of this section we provide details about items we consider at each level, tests and input values used for absolute evaluation and indicators that may contribute to their differential evaluation when direct input is not available.
The identification of relevant items and indicators, as well as their classification has been made possible thanks to the strict interaction with a team of clinicians in geriatrics from the ``Ospedale S. Gerardo'' in Monza (Milano, Italy). Under their supervision, we analyzed their protocols in elderly care and formalized our domain knowledge accordingly.

\begin{description}
\item[State]
This class of items includes:

		\begin{enumerate} 
          \item comorbidity, intended as the simultaneous presence of two (or more) chronic diseases or conditions in a patient;
          \item number of different classes of drugs;
		  \item quality of the environment, computed from average lighting, humidity and temperature;
		  \item Body Mass Index (BMI) value.
        \end{enumerate}
The evaluation of items at the state level can depend on specific input from caregivers.

\item [Functionalities]
Items in this class include disabilities in the following aspects of the person's health state:

\begin{enumerate}
  \item \emph{balance} and \emph{gait}, initially evaluated through the Tinetti-POMA~\cite{poma86} medical scale; indicators are represented by i) assessment of related pathologies, ii) drug intake and iii) indicators representing a few aspects of the scale that can be captured by the wearable sensor and evaluated through ad-hoc feature aggregation: standing, sitting, laying, turning and walking; 
  \item \emph{nutrition}, initially evaluated by means of the Mini Nutritional Assessment~\cite{mna94} test; the indicator is the Body Mass Index (BMI), stored every hour;
  \item \emph{vision}, initially evaluated through specific tests; indicators are the average levels of light of an area when the person is in that area, evaluated and stored every hour;
  \item \emph{hearing}, initially evaluated through specific tests; indicators are the response time to a ringing bell, evaluated and stored periodically as a result of a test performed under specific conditions;
   \item \emph{cognition}, evaluated by means of the Mini Mental~\cite{mmt75} and Clock Drawing~\cite{cdt04} tests; indicators are represented by test results;
  \item \emph{insomnia}, initially evaluated by means of a questionnaire; the indicator is the quality of sleep in three different moments of the sleeping period: start of the sleeping period, heart of the night, early morning;
  \item \emph{emotional stability}, initially evaluated by means of the GDS test~\cite{gds83}; the indicator is a computer-aided version of the GDS test. 
\end{enumerate}

\item [ADLs]
For items in this class, we are interested in the estimated level of dependency in performing the Activities of Daily Living (ADLs) mentioned in the Katz scale~\cite{Katz70}, in particular:

\begin{enumerate}
  \item \emph{mobility} initially evaluated through the PASE scale; indicators are the walking speed, the number of steps and the average daily walking time;
  \item \emph{dress} evaluated according to specific tests performed by caregivers;
  \item \emph{eat} evaluated by means of the Mini Nutritional Assessment Test~\cite{mna94};
  \item bathing has no indicators in the current version;
  \item toileting has no indicators in the current version.
\end{enumerate}

We want to point out that reasoning with ADLs is not aimed at activity recognition as in other approaches to monitoring~\cite{Pollack05}. We start from absolute evaluations as input and then concentrate on possible inter-dependencies that may influence the level of autonomy with which a person performs an ADL, thus predicting a risky evolution of the health state according to correlations with other items as detailed in Section~\ref{sec:reasoning}. We believe that the analysis of such dependencies is crucial for prevention. 

Instrumental Activities of Daily Living (IADL) from the Lawton scale~\cite{Law88} have not been included either, except for drugs intake and use of the telephone. This choice has been guided by the fact that they have a lower impact on other health-related items and the evaluation is too complex to be performed in a non intrusive way.

\item [Risk Assessment]

The risks we consider in the SINDI system are represented by the potentially most dangerous situations for elderly people at home, namely:
\begin{enumerate}
  \item \emph{falls}, the risk is initially evaluated according to the Tinetti POMA scale; it can be influenced by specific pathologies and drug intake used as indicators;
  \item \emph{depression}, the risk is initially evaluated according to the GDS scale~\cite{gds83};  it can be influenced by specific pathologies and drug intake used as indicators;
  \item \emph{frailty}, initially evaluated through a combination of GDS test, Mini Mental test and Katz evaluation; indicators are related to some profile information such as walk speed and age, but it is very complex to evaluate frailty without the support of a caregiver; thus the evaluation based on indicators only is to be supported by additional reasoning as detailed in Section~\ref{sub:health}.
  \item \emph{isolation}: initially evaluated through the GDS test; the indicators are the level of social activity (number of contacts, time spent out of the house) and the presence of affective disorders.
\end{enumerate}

\end{description}

In order to make it possible to evaluate indicators and items in a context-aware fashion, SINDI's knolwedge model of health consists of a two-layered graph: a layer where nodes are indicators and a layer where nodes are items

\begin{figure}[t]
\includegraphics[width=0.99\textwidth,keepaspectratio]{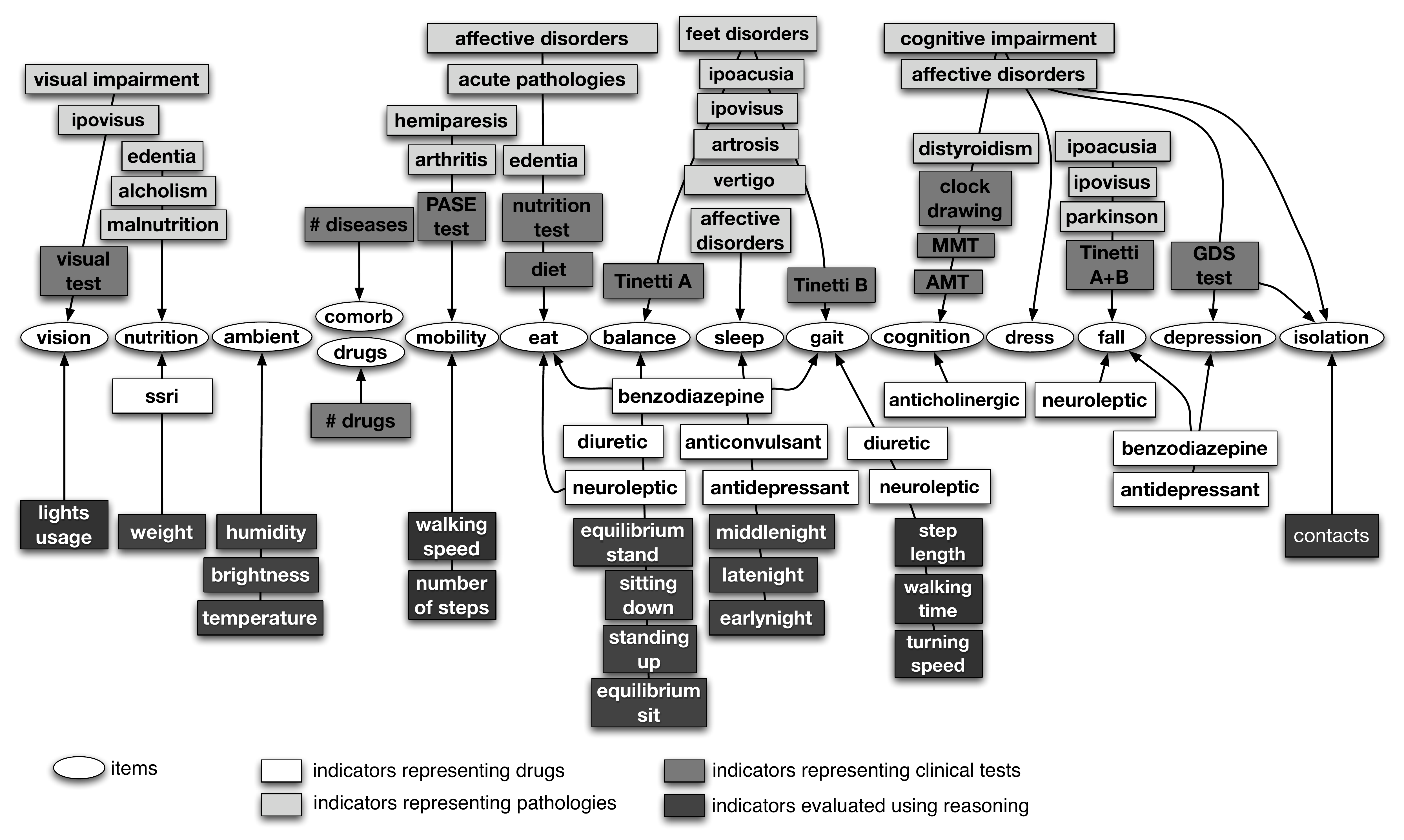}
\caption{SINDI's knowledge model of health: influences between indicators and items}\label{fig:model1}
\end{figure}

\begin{figure}[t]
\includegraphics[width=0.80\textwidth,keepaspectratio]{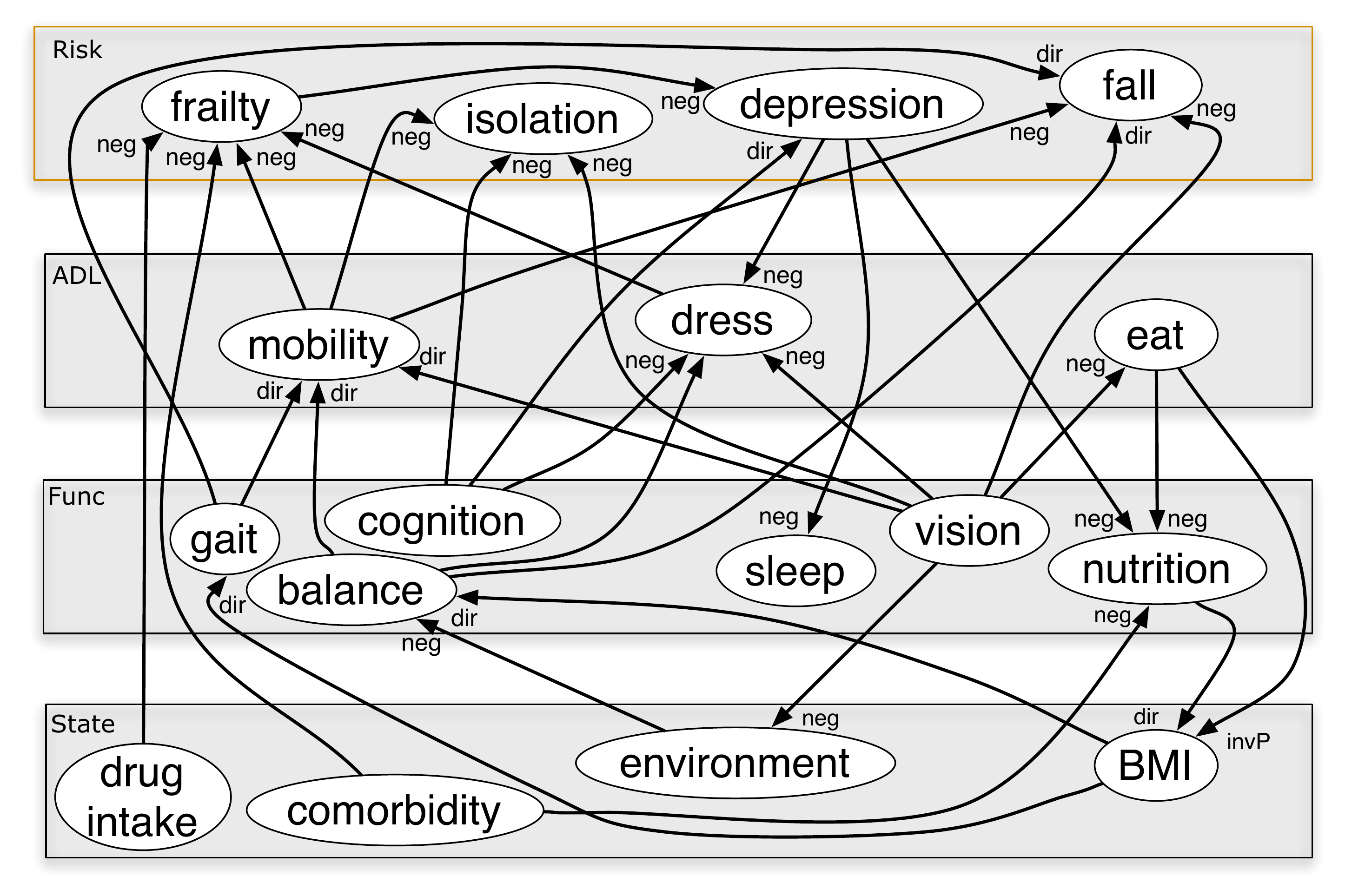}
\caption{SINDI's knowledge model of health: influences among items}\label{fig:model2}
\end{figure}

Indicator nodes can be connected to item nodes through arcs that represent dependencies between the differential evaluation of the indicators and the differential evaluation of the related item.
One item node $I_1$ can be connected to another item node $I_2$ through a dependency arc that represents the influence of the differential evaluation of $I_1$ on the differential evaluation of $I_2$.
The two layers of the graph-based representation of SINDI's knowledge model of health are illustrated in Figures~\ref{fig:model1} and~\ref{fig:model2}.


SINDI's knowledge model of health allows different kinds of dependencies: 

\begin{enumerate}
  \item {\bf neg/pos}: strictly negative/positive direct influence of the evaluation of a source node on the evaluation of a target node;
  \item {\bf invN/invP}: strictly negative/positive inverse influence of the evaluation of a source on the evaluation of a target node;
  \item {\bf dir/inv}: directly/inversely proportional influence of the evaluation of a source node on the evaluation of a target node;
\end{enumerate} 

Dependencies can be specified by caregivers and are automatically mapped into ASP as follows:

\begin{equation}\label{eq:dependency}
\begin{array}{ll}
link(Type,Ind,I) & to\ express\ influence\ of\ indicator\ Ind\ on\ item\ I\\
influence(Type,I_1,I_2) & to\ express\ influence\ of\ item\ I_1\ on\ item\ I_2\\
\end{array}
\end{equation}

where Type $\in$ \{pos, neg, invP, invN, dir, inv\}.

Since we believe feedback represents the key for effective preventive interventions, an additional part of the domain knowledge of SINDI is related to the representation of admissible feedback from the system to the person monitored.
Furthermore, given that each person has a different clinical history of cognitive decline and reacts in different ways to external stimuli, it is extremely important to select the most appropriate feedback according to the context.

\begin{table}
\caption{Classification of feedback outputs for fall prevention.}
\label{tab:feedback}
\begin{center}
{\footnotesize
\begin{tabular}{llll}\hline
Class & Output & Form & System Action \\
\hline \hline
Environment & do not walk in the dark & S, A, N & turn lights on \\
 & check temperature & S, A, N & vocal warning\\
 & incoming call & S, A, AA, N  & blinking lights \\
 & remove clutter & S, A, N  & vocal warning\\
 & emergency & AA  & call 911 \\
Behaviour & keep active & S, A, N, R  & propose exercises\\
 & stand up slowly & S, A, N  & vocal warning \\
 & make sure carpets are fixed & S  & vocal warning \\
 & use chair to get dressed & S  & vocal warning \\
Clinical actions & review drugs & S  & on screen \\
 & see a specialist & S, R  & on screen \\
\hline

\end{tabular}
}
\end{center}
{\footnotesize
\begin{tabular}{lllll}
(S) suggestion & (AA) alarm & (A) alert & (N) notification & (R) reminder \\
\\
\end{tabular}
}
\end{table}

The system can provide feedback in five different ways: 

\begin{itemize}
\item \emph{suggestions} according to the medical practice and the results of the prediction task;
\item \emph{alerts} when the system identifies behaviors or situations that are potentially dangerous according to the results of the prediction task;
\item \emph{alarms} when specific environmental or clinical conditions are detected;
\item \emph{notifications} when the system receives new input or terminates the inference process;
\item \emph{reminders} according to an agenda. 
\end{itemize}

The main difference between a suggestion and an alert is that the second is triggered by the identification of a specific behavior and may generate an immediate action as an output (e.g. a blinking light to indicate that there is a call), while the first is purely based on the medical knowledge encoded in the system and gives a report as output.
Alarms also generate an action but unless alerts that are generated as a result of a prediction (potential risk), they are raised as a result of an evaluation (effective risk), thus they usually need a more urgent response (e.g. a call to the caregiver when a fall is detected). 
In our first specification reminders do not include support on how to perform complex activities as in~\cite{Autominder03,Coach06}.
In the current release, reasoning aimed at prevention is at its early stage, and the system deals only with simple reminders according to an agenda. We are aware of the fact that logic programming techniques are promising also to solve planning problems, and we want to further investigate the potential of our formalism in guiding the person in the correct execution of daily activities. However, this is not part of the main objectives of the SINDI system, where we focus on the interpretation of incomplete sensor data aimed at predicting health evolutions exploiting a graph-based computational model of health dependencies.

Independently of how it is delivered, a feedback action can be related to: 
        \begin{itemize}
          \item the environment: making the environment safer and of better quality, improving interaction with the environment, e.g. a phone call that is not acknowledged by the patient can trigger actions like reducing the volume of the TV or blinking a light;
          \item the user's behavior: suggesting how to modify habits when the health assessment indicates risky conditions or providing reminders according to an agenda;
          \item the clinical setting: consulting a doctor, suggesting a more accurate test, reviewing a therapy, reminding medical appointments, and so on.
        \end{itemize}

The combination of the results of the prediction task, domain knowledge and context-related knowledge about the person and the environment is used to determine i) \emph{what} should be provided as feedback, ii) in \emph{which form} and iii) \emph{when}.
In the actual implementation of SINDI, we focused on feedback outputs related to prevention of falls.
The contents of the feedback (later referred to as feedback outputs) are identified according to the evidence-based medical knowledge derived from experimental trials and encoded as logic facts~\cite{Con97,Stuck99,Rub06,Nes03}.

The most appropriate form of feedback is inferred by the system on the basis of the results of reasoning: the same feedback can be provided in different forms and at different times (see Section~\ref{sub:health} for details).

A feedback can be provided as soon as it is inferred or at a later time. Alarms are usually immediate, while other forms of feedback can be performed immediately or at a later time according to:
 	\begin{itemize}
              \item triggers: pushing a button at a specific time or when particular conditions hold;
              \item user/caregiver preferences: qualitative ordering to identify more urgent/important suggestions according to the environmental/personal context setting and the form of feedback;
              \item static ordering: certain forms of feedback may have higher priority than others, simply because of their nature; similarly, some communication patterns can be preferred to others on the basis of clinical settings.
	\end{itemize}

As an example, consider the classification of feedback outputs for fall prevention illustrated in Table~\ref{tab:feedback}~\cite{AC-workshop09}.
Knowledge contained in Table~\ref{tab:feedback} is represented as logic predicates of the form  $possible\_form(Output,Form)$ associating the content of a feedback to its possible forms.

Knowledge about system's reactions is used for prevention as illustrated in Section~\ref{sub:health} and it is mapped into logic predicates as follows:
\begin{itemize}
 \item for each possible form $f_i\in F=\{s, a, aa, n, r\}$ for an output $X$ use predicate $possible\_form(X,f_i)$;
 \item output $X$ triggered by some events in the form $f_i \in F$ is represented by predicate $feedback\_form(X,f_i)$;
 \item action $Z$, consequence of output $O$ triggered in form $F$ is represented by predicate $do\_action(O,F,Z)$;
 \item the action of prompting output $O$ triggered in form $F$ through channel $C=\{audio,video\}$ at time $T=\{immediate, endOfDay\}$ is represented by predicate $do\_prompt(O,F,C,T)$.
 \item actions observed by the system are represented by predicate $action\_observed(A,T)$ where $T$ represents a discrete time stamp and $A$ is an action among those that can be recognised by the reasoning process.
\end{itemize}

\section{Reasoning Support for Intelligent Monitoring}\label{sec:reasoning}

We refer to an \emph{intelligent monitoring system} as a monitoring system that is able to support understanding and decision through the interpretation of data according to an appropriate model of the domain.

As introduced in previous sections, SINDI's inference for intelligent monitoring is based on context interpretation, i.e. the process of reasoning about context-dependent sensor data through specific inference rules in order to provide a consistent view of the world.
In pervasive environments, context-dependent data can arise from different sources; for example data may be gathered by sensors or collected from several knowledge-bases. The incompleteness and heterogeneous nature of such data stress the need for expressive reasoning techniques in order to implement effective, context-aware reasoning.

We already identified three classes of reasoning tasks performed by the SINDI system: Context Interpretation, Health Assessment (evaluation) which uses results of Context Interpretation and Health Evolution (prediction, explanation, reaction) based on results of Health Assessment and SINDI's model of health.

In the following subsections we illustrate SINDI's reasoning tasks for each class.

\subsection{Context Interpretation}\label{sub:ctx-reasoning}

Data gathered by the sensors may be noisy even after aggregation, but their combination may yield a reliable interpretation. The expressive power of ASP is used to disambiguate unclear situations (e.g., where the person is) by combining heterogeneous data sources and using defaults, nondeterministic choice and preferences to select the best candidates in the space of the solutions.

The continuous measurements provided by sensors are stored in the database of SINDI (Fig.~\ref{fig:layers}). In the actual implementation, tasks related to context interpretation are performed every hour. Discrete time is in seconds and sensor data are opportunely aggregated and provided when values change beyond a given threshold. As an example, the temperature is aggregated as the average temperature of a room and a new logic fact is generated whenever the temperature value changes of at least 1 degree Celsius. We are currently working on a wrapper that will make it possible to use the Clasp solver as a permanently running API that we can feed with aggregated sensor data as soon as they are available. This will make the reasoning process faster.

In order to understand how reasoning helps in the interpretation of imprecise sensor data, let us consider localization of the person in a position $X,Y$ of the grid representing the home environment.
SINDI's localization component is based on the intensity variations of the radio signals exchanged between nodes, filtered by a bayesian filter. Unfortunately, it is not always true that the higher the measured intensity of a signal from a node, the closest the person is to that node. 

Given proximity values with a certain accuracy $P$ at a given time $T$ (provided as facts of the form $in(X,Y,T,P)$), the ASP program takes available sensor data that can be used to validate proximity signals and to reason about several preference criteria to identify the best solutions.


All possible positions are generated for a given time $T$ and the optimal solutions are obtained by applying different combination of the preference criteria: when the proximity signals are available, the highest signal which is most coherent and best move is preferred. If no such position exists, then coherence is preferred to the best move criterion; finally we give up coherence if no such position exists.
When the proximity signal is not available, the same principle applies, except that we maximize the support obtained form other sensor data (namely range finder and infrared) for a given location $L$ of the grid, rahter than maximizing the signal strength. Sensor data are represented as facts of the form $sensed(S,L,T)$, where $L$ represents the location $loc(X,Y)$, $S$=distance for the range finder and $S$=motion for the infrared.

The correspondent encoding is as follows:

\begin{small}
\begin{verbatim}
time(1..207).
has_data(T)       :- sense(S,X,Y,T).
has_rssi(T)       :- in(X,Y,T,P).
invalid(loc(X,Y)) :- wall(X,Y).
sensed(rssi,P,loc(X,Y),T) :- in(X,Y,T,P).
sensed(S,loc(X,Y),T)      :- sense(S,X,Y,T).

%%%%%%%%%%%%%%%%%%%%%%%%%%%%%%%%
% Best Move
%%%%%%%%%%%%%%%%%%%%%%%%%%%%%%%%
location(L,T) :- sensed(rssi,P,L,T).
location(L,T) :- sensed(S,L,T).
location(L)   :- location(L,T), not invalid(L).
length(L,T)   :- location(N;M), dist(N,M,L,T).

dist(loc(X,Y),loc(U,V),#abs(X-U) + #abs(Y-V),T) :- location(loc(X,Y),T), location(loc(U,V),T-1).
distance(D,L,T)    :- location(L,T), at(M,T-1), dist(L,M,D,T), location(L).
best_distance(D,T) :- D = #min [distance(E,L,T) : location(L,T): length(E,T): E>0 = E ],
                      time(T;T-1).
best_movement(L,T) :- distance(D,L,T), best_distance(D,T).

%%%%%%%%%%%%%%%%%%%%%%%%%%%%%%%%
% Best Coherence
%%%%%%%%%%%%%%%%%%%%%%%%%%%%%%%%
is_passage(X,Y)           :- passage(X,Y,R1,A1,R2,A2).
is_cell(X,Y)              :- cell(X,Y,R,A).
data_expected(S,loc(X,Y)) :- data_ex(S,X,Y).
ex_support(N,L)    :- N = [data_expected(S,L) : sensed_type(S)], location(L,T).
p_support(L,T,C)   :- C = [sensed(S,L,T) : data_expected(S,L)], location(L,T), not invalid(L).
coherence(L,T,P)   :- p_support(L,T,C), P=(100*C)/N, ex_support(N,L), C>0, N>0.
coherence(L,T,C)   :- p_support(L,T,C), C=0, ex_support(N,L), N>0.
coherence(L,T,100) :- p_support(L,T,C), ex_support(N,L), C=0, N=0.
most_coherent_time(T,M) :- M = #max [coherence(L,T,C) : coherence(L,T,C) = C], 
                           time(T), perc(C), M>=0.
most_coherent(L,T)      :- most_coherent_time(T,C), coherence(L,T,C).
best_coherence(L,T)     :- most_coherent(L,T).

%%%%%%%%%%%%%%%%%%%%%%%%%%%%%%%%
% Multicriteria optimization
%%%%%%%%%%%%%%%%%%%%%%%%%%%%%%%%
criterion(1..4).
criterion(1,L,T) :- location(L,T), best_movement(L,T), best_coherence(L,T).
criterion(2,L,T) :- location(L,T), best_movement(L,T).
criterion(3,L,T) :- location(L,T), best_coherence(L,T).
criterion(4,L,T) :- location(L,T).

%%%%%%%%%%%%%%%%%%%%%%%%%%%%%%%%
% Optimization using criterias
%%%%%%%%%%%%%%%%%%%%%%%%%%%%%%%%
signal(P)          :- sensed(rssi,P,L,T).
signal_data(N)     :- count_s(N,L,T).
count_s(N,L,T)     :- N = [sensed(S,L,T) : sensed_type(S)=1], location(L,T), 
                      not has_rssi(T). 
reward(L,T,N)      :- not has_rssi(T), has_data(T), count_s(N,L,T).
sense_power(P,T,C) :- sensed(rssi,P,L,T), criterion(C,L,T).
sense_num(N,T,C)   :- count_s(N,L,T), sensed(S,L,T), criterion(C,L,T).
best_value(C,B,T)  :- criterion(C), time(T), signal(B), 
                      B = #max [sense_power(P,T,C):signal(P)=P],
                      not best_value_p(D,T) : criterion(D): D<C.
best_value(C,B,T)  :- criterion(C), time(T), signal_data(N), B>=0,
                      B = #max [sense_num(N,T,C):signal_data(N) = N],
                      not best_value_p(D,T) : criterion(D): D<C.
best_value_p(C,T)  :- best_value(C,B,T).
best_location(L,T) :- best_value(C,B,T), sensed(rssi,B,L,T).
best_location(L,T) :- best_value_p(C,T), sensed(S,L,T), not has_rssi(T).

%%%%%%%%%%%%%%%%%%%%%%%%%%%%%%%%
% Generate and Test
%%%%%%%%%%%%%%%%%%%%%%%%%%%%%%%%
1 { at(L,T) : location(L) } 1 :- time(T).
:- at(L,T), not best_location(L,T).

#hide.
#show at/2.
\end{verbatim}
\end{small}

Besides localization, context interpretation is in charge of understanding basic beaviours that may be important for health assessment, such as the night activity. In the following subsection we illustrate how this is obtained and used in the following reasoning step.

\subsection{Context-aware Health Evaluation}\label{sub:eval}


At each reasoning cycle, results of the context interpretation process are used to infer consistent evaluations of indicators and items, combining specific logic rules with medical inputs and results of data aggregation.
The evaluation process produces both an absolute evaluation and a differential evaluation (with respect to results of the previous inference cycle).
Admissible values for each indicator and item are part of the medical knowledge and are encoded in the system, while their differential evaluation has four possible outcomes: worsening, improvement, no substantial change, undefined.


In general, absolute evaluation of items is available only as inputs from caregivers according to results of a specific tests. As for indicators, their absolute evaluations can be based on i) results of specific evaluation by clinicians (e.g. hearing functionalities), ii) results of data aggregation (e.g. quality of movement) and iii) results of ad-hoc logic rules (e.g. quality of sleep).

Absolute evaluation is represented in form of logic predicates of the form:
\begin{equation}\label{eq:dependency}
\begin{array}{ll}
obsInd(Ind,V,H) & for\ indicators\\
obsItem(I,V,H) & for\ items\\
\end{array}
\end{equation}

where H is a time stamp identifying the hour associated to the inference cycle the evaluation refers to\footnote{Timestamp $H=0$ is associated the hour being evaluated, while timestamps $H>0$ refers to previous inference steps: the highest $H$, the oldest the hour.}, and value $V$ is defined over specific ranges in the knowledge base. The current hour being processed is provided by fact $hour(N), N=0..23$.

In order to highlight the advantages of using reasoning in the evaluation process, we provide an example of how inference can help in this reasoning phase.

\begin{example}
Consider the indicator \emph{quality of sleep} which is one of the most interesting in assessing the well being of the elderly because it can be used as a predictor of worsening conditions.
Nonmonotonic reasoning based on ASP makes it possible to combine several context-dependent informations inferred by the context interpretation process to determine a consistent evaluation of the overall quality of sleep. A simplified version of the encoding used to connect consecutive reasoning cycles to determine night activity is as follows:

\begin{small}
\begin{verbatim}
night    :- hour(N), N<8.
night    :- hour(N), N>21.
awake(T) :- not in_bed(T), time(T), localized(T), night.

sleep_interrupt(T) :- in_bed(T1), awake(T), T1<T. 
sleep_interrupt(T) :- obsInd(S,ok,1), link(L,S,sleep), awake(T), time(T). 
sleep_interrupt(T) :- obsInd(S,moderate,1), link(L,S,sleep), awake(T), time(T). 

back_to_bed(T) :- awake(T1), in_bed(T), T1<T, obsInd(S,mild,1), link(L,S,sleep). 
back_to_bed(T) :- sleep_interrupt(T0), awake(T1), in_bed(T), T0<T1, T1<T, 
                  link(L,S,sleep), time(T0;T1;T). 
bad_sleep(T)   :- in_bed(T), attribute_obj(loadVolatility,bed,N,T), N!=stable.

poss_early_awake(T) :- sleep_interrupt(T), not in_bed(T1), T<=T1, time(T1).
poss_early_awake(T) :- awake(T), obsInd(S,mild,1), link(L,S,sleep), time(T).
n_early_awake(T)    :- poss_early_awake(T), back_to_bed(T1), time(T1), T<T1. 
\end{verbatim}
\end{small}

The following piece of code is in charge of evaluating indicators associated to the quality of sleep:

\begin{small}
\begin{verbatim}
period(N,earlynight)  :- hour(N), N>21. 
period(N,middlenight) :- hour(N), N<2. 
period(N,latenight)   :- hour(N), 2<=N, N<5.

obsInd(S,ok,0)   :- link(L,S,sleep), not sleep_interrupt(T1), time(T1), 
                    in_bed(T), T1!=T, obsInd(S,ok,1), period(N,S), hour(N), 
                    not obsInd(S,mild,0), not obsInd(S,moderate,0), 
                    not obsInd(S,consistent,0).
obsInd(S,mild,0) :- link(L,S,sleep), period(N,S), bad_sleep(T), time(T), 
                    not obsInd(S,moderate,0), not obsInd(S,consistent,0).
obsInd(S,moderate,0)   :- link(L,S,sleep), period(N,S), back_to_bed(T),  
                          time(T), not obsInd(S,cosistent,0).
obsInd(S,consistent,0) :- link(L,S,sleep), period(N,S), poss_early_awake(T1), 
                          not n_early_awake(T1), time(T).
\end{verbatim}
\end{small}

\end{example}

Differential evaluations are obtained, when possible, as a measure of the value increase or decrease derived by comparing the values of the current inference cycle with values at the previous inference cycle.

If we imagine to assign a color to each node according to the differential value (assigning no color to undefined values) the evaluation process returns a partially labelled graph as output. At this point, the SINDI reasoning component can further help in the differential evaluation of items.
In fact, differential evaluation of an item $I$ is not straightforward when no direct evaluation is available for $I$. When this is the case, we can still use reasoning to infer a coherent differential evaluation according to i) differential evaluations of all indicators $Ind_i$ that may influence $I$ and ii) multiple dependencies between each $Ind_i$ and $I$.

The effect of a single dependency arc from a source $Ind$ to a target $I$ according to the type of arc dependency, is summarized in Table~\ref{tab:influence}. The sign associated with a dependency arc connecting indicator $Ind$ and item $I$ is determined according to the differential evaluation of $Ind$ and the label of the arc.

\begin{table}
\caption{Influence of arc from Ind to I according to arc label}
\begin{center}
\begin{tabular}{c|cccccc}
\textbf{Sign of Source Node} & \textbf{pos} & \textbf{neg} & \textbf{invP} & \textbf{invN} & \textbf{dir} & \textbf{inv} \\
\hline
+ & + & ? & -- & ? & + & -- \\
-- & ? & -- & ? & + & -- & + \\
= & = & = & = & = & = & = \\
? & ? & ? & ? & ? & ? & ? \\
\end{tabular}
\end{center}
\label{tab:influence}
\end{table}

Given that different kinds of dependencies are allowed in SINDI's knowledge model, and they are potentially contradictory or incomplete, we need an algorithm to combine the effects of the combination of such influences of several indicators $Ind_i$ on item $I$ so as to provide a coherent differential evaluation for $I$.

\begin{description}
 \item[Evaluation process:]
 \item [a)] If no differential evaluation is available for item $I$, $I$ should be evaluated by using dependencies
 \item [b)] Influences determining a ``+'' sign prevail if there are no influences giving signs ``--'' or ``?''
 \item [c)] Influences determining a ``-'' sign prevail if there are no influences giving signs ``+'' or ``?''
 \item [d)] Influences determining a ``='' sign prevail if it is the only sing produced by all other dependencies 
 \item [e)] If no differential evaluation is produced for item $I$, further reasoning is needed to guess its possible evolution
\end{description}

We represent an absolute (available as an input) differential evaluation of item $I$ with predicate $diff\_item(L,I,Val,T)$; an inferred differential evaluation of item $I$ is computed by the reasoning engine according to values of indicators $Ind_i$ connected to $I$, and it is represented by the predicate $diff\_item\_inferred(L,I,Val,T)$.

The sign of the dependency is represented by predicate $infl\_sign(I,Ind,S,T)$ with $S\in \{0, -1, 1\}$.  The sign $S$ of $infl\_sign(I,Ind,S,T))$ depends on the differential evaluation of $Ind$ represented by predicate $diff\_ind(Ind,Val,T)$ and the type of link $link(Type,Ind,I)$ as illustrated in Table~\ref{tab:influence}. Differential values $Val$ correspond to a sign $S$ as follows: \emph{better} corresponds to ``$+$'', \emph{worse} corresponds to ``$-$'' and \emph{equal} corresponds to ``$=$''. For sake of simplicity, in the knowledge representation framework we represent signs as numbers so that they can be multiplied by their weights for optimization: $0$ represents no changes, $-1$ represents worsening conditions and $+1$ represents improvements.

The logic rules implementing this behaviour would look like the following:

\begin{small}
\begin{verbatim}
to_evaluate(L,I) :- not diff_item(L,I,Val,T).
n_invariant(I)   :- infl_sign(I,Ind,S,T), S!=0.

diff_item_inferred(L,I,S,T) :- to_evaluate(L,I), infl_sign(I,Ind,S,T), 
                               not infl_sign(I,Ind1,S1,T), S1=S*(-1).

diff_item_inferred(L,I,0,T) :- to_evaluate(L,I), infl_sign(I,Ind,0,T), 
                               not n_invariant(I).

labeled(I)    :- diff_item_inferred(L,I,S,T).
to_guess(L,I) :- to_evaluate(L,I), not labeled(L,I).

\end{verbatim}
\end{small}

Intuitively, this reasoning process could provide additional differential evaluation useful for prediction by the higher level reasoning tasks. 
We could make this process more precise by using weighted arcs to represent influences of indicators $Ind_i$ on item $I$; in this way we could label item $I$ with the sign that has the higher sum of weights. The major problem of this approach is that in the healthcare domain, the impact of a medical dependency is not static, thus weights should be computed dynamically according to the whole clinical situation. This aspect is not even clear to caregivers and we plan to investigate this issues with domain experts to find a viable solution. 

We show a graphical example that illustrates the results of reasoning in a simple two-layered graph.

\begin{example}
Consider the dependency graph illustrated in Figure~\ref{fig:ex-eval}.

\begin{figure}[h]
\includegraphics[width=0.85\textwidth,keepaspectratio]{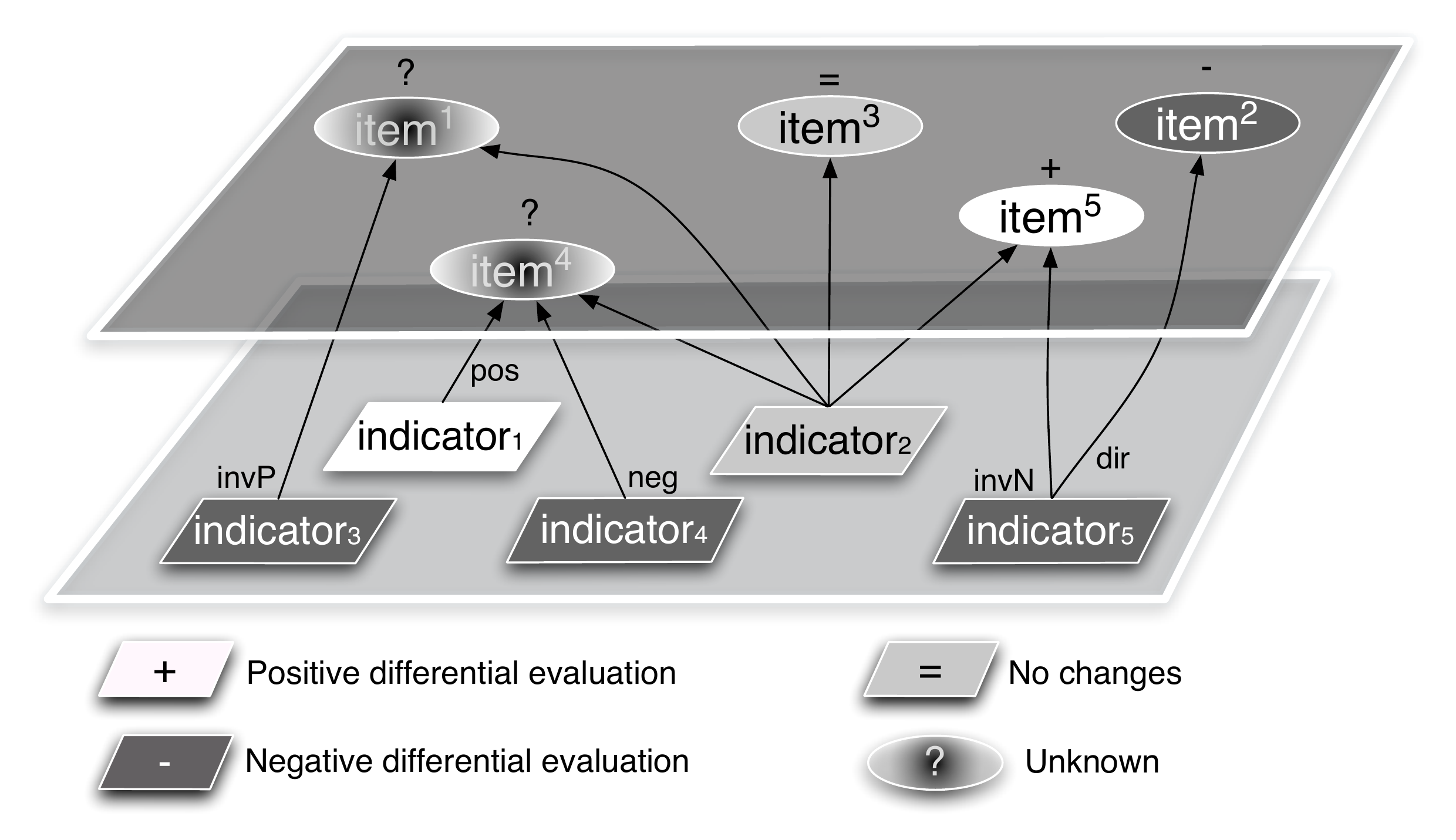}
\caption{Results of the evaluation reasoning process for a simple dependency graph}\label{fig:ex-eval}
\end{figure}

Note that the label of arcs connecting $Indicator_2$ with items $Item_i,\ i=\{1,3,4,5\}$ is not important for their differential evaluation. In fact $Indicator_2$ is marked as stable (its sign is ``$=$'') thus its influence is the same according to Table~\ref{tab:influence}.

\end{example}

Often it can be the case that the output of the evaluation phase is a partial labeling of the graph of items and related dependencies.
The following reasoning process of SINDI starts from this incomplete information about health in the Health Assessment layer and uses the computational power of the ASP framework in order to:

\begin{itemize}
  \item predict possible evolutions in terms of differential evaluations of items that have not been labeled as a result of the evaluation phase and provide a qualitative analysis of results;
  \item investigate the plausible causes of predicted and inferred health changes according to the dependency graph;
  \item use results of prediction to support caregivers in planning appropriate interventions.
\end{itemize}

These reasoning tasks will be detailed in Section~\ref{sec:reasoning}.
 
\subsection{Health Evolution}\label{sub:health}

According to what has been described in Section~\ref{sub:model}, the Health Assessment layer of the graph representing SINDI's logic-based model of health includes a well-defined set of items and correlations among them. The correspondent graph is partially labeled with the results of the evaluation process.

In this section we want to focus on how the system reasons about this incomplete and potentially incoherent information about item values to predict health changes and suggest appropriate interventions.


With the introduction of reasoning in health monitoring, we want to address the fact that caregivers need to be supported in understanding patients' physical, mental and social settings as they evolve, by i) predicting what could follow with respect to particular changes in one or more aspects of the patients' general health, ii) identifying correlated aspects that may be the cause for a predicted change in the patient's general health, iii) giving appropriate (context-dependent) feedback to the patient to educate him to correct behavior and collecting his reactions to this feedback. 

The first aspect refers to prediction, i.e. the identification of plausible effects of certain changes in items' values on values of unlabeled items; intuitively, this is done by considering all possible consistent values for the missing information according to SINDI's logic model of dependencies between items; prediction makes it possible to act \emph{before} major symptoms and to plan appropriate short- and long-term interventions, thus reducing risks.

The second aspect may look similar to diagnosis, but we prefer to call it \emph{Local Explanation} in that it gives reasons for the differential evaluations of an item $I$ provided as a result of the prediction task, when the inferred sign for $I$ is the same in all solutions.
It is a \emph{local} process rather than a case-based one, in that it takes into account results of reasoning under particular clinical and environmental conditions.

The third aspect is related to the identification of those interventions (provided as direct actions performed by the system or as suggestions) that may keep health changes within safe boundaries.

In the remaining part of this section we give implementation details for each of these classes of tasks.

\begin{description}
\item[ ]
\item[Prediction]


In order to determine all total consistent labelings of the graph resulting from the evaluation process, the inference mechanism takes into account unlabeled items and generates consistent labeling.
The instance provided as input to the ASP program includes sign $S$ for item $I$ represented by predicate $obs\_label(I,S)$ or $inferred\_label(I,S)$ provided by the evaluation process, and unlabeled items $to\_guess(I_k)$.

Labels are obtained according to the following procedure:

\begin{equation}
\begin{array}{lll}

 \forall\ I\ | & \exists\ obs\_label(I,S): & \\
 & compute\ sign(I,I_k,S') & for\ each\ arc(I,I_k)\ as\ in\ Table~\ref{tab:influence};\\
 & compute\ weight\ W(I,I_k)\ | & W(I,I_k)=5\ if\ \exists\ obs\_label(I,S), \\
 & & W(I,I_k)=1\ if\ \exists\ inferred\_label(I,S);\\
 \forall\ I_k\ | & to\_guess(I_k): & \\
 & compute\ total\ weight & W_{tot}(I,I_k,S')=\Sigma_{k=1}^n\ W(I,I_k)\ |\ sign(I,I_k,S');\\
 & compute\ winning\ sign\ S_w & ilab(I_k,S_w).\\
\end{array}
\end{equation}

As a result of the prediction task, the ASP logic program may yield different solutions from which we extract:

\begin{enumerate}
  \item guessed signs $S'$ for item $I$ that are true in all possible solutions, represented by predicate $ilab(I,S')$,
  \item for all other items $I_g$ for which there are several possible guesses, we prefer the solution where each $I_g$ is labeled with sign $S_g$ such that the sum $\Sigma_{g=1}^n\ W_{tot}(I_g, S_g)$ is maximized.
\end{enumerate}
 
The following example shows results of prediction task on a small graph that has been partially labeled by the evaluation process.

\begin{example}\label{ex:pred}
Consider results of the evaluation process as in the first graph of Table~\ref{tab:ex-pred}. 
Two direct observations are available: $obs\_label(item_3,-1)$ and $obs\_label(item_2,0)$. We set the weight of each arc from $I_j$ to $I_h$ to $5$ if sign of $I_j$ is provided by the evaluation process, to $1$ otherwise.
Possible consistent guesses are represented by three answer sets containing, among others, the following predicates:

{\small
\begin{displaymath}
\begin{array}{lll}
S_1 = & \{ & count\_infl(item_4,-1,5),\ count\_infl(item_4,0,1),\ count\_infl(item_5,0,5),\\
& & count\_infl(item_2,-1,5),\ count\_infl(item_2,0,2), ilab(item_6,0), \\
& & ilab(item_3,-1),\ ilab(item_4,-1),\ ilab(item_5,0),\ ilab(item_2,-1),\ ilab(item_1,0) \}\\
S_2 = & \{ & count\_infl(item_4,-1,1),\ count\_infl(item_4,-1,5),\ count\_infl(item_5,0,5),\\ 
& & count\_infl(item_2,-1,5),\ count\_infl(item_2,0,1),\ ilab(item_6,0), \\ 
& & ilab(item_3,-1),\ ilab(item_4,-1),\ ilab(item_5,0),\ ilab(item_2,-1),\ ilab(item_1,-1)\} \\
S_3 = & \{ & count\_infl(item_4,-1,5),\ count\_infl(item_5,0,5),\ count\_infl(item_2,-1,5),\\
& & count\_infl(item_2,0,1),\ count\_infl(item_2,1,1), ilab(item_6,0),\\
& & ilab(item_3,-1),\ ilab(item_4,-1),\ ilab(item_5,0),\ ilab(item_2,-1),\ ilab(item_1,1) \} \\
\end{array}
\end{displaymath}
}

\begin{table}
\caption{Graph for Example~\ref{ex:pred}}
\begin{center}
\begin{tabular}{cc}
\includegraphics[width=0.50\textwidth,keepaspectratio]{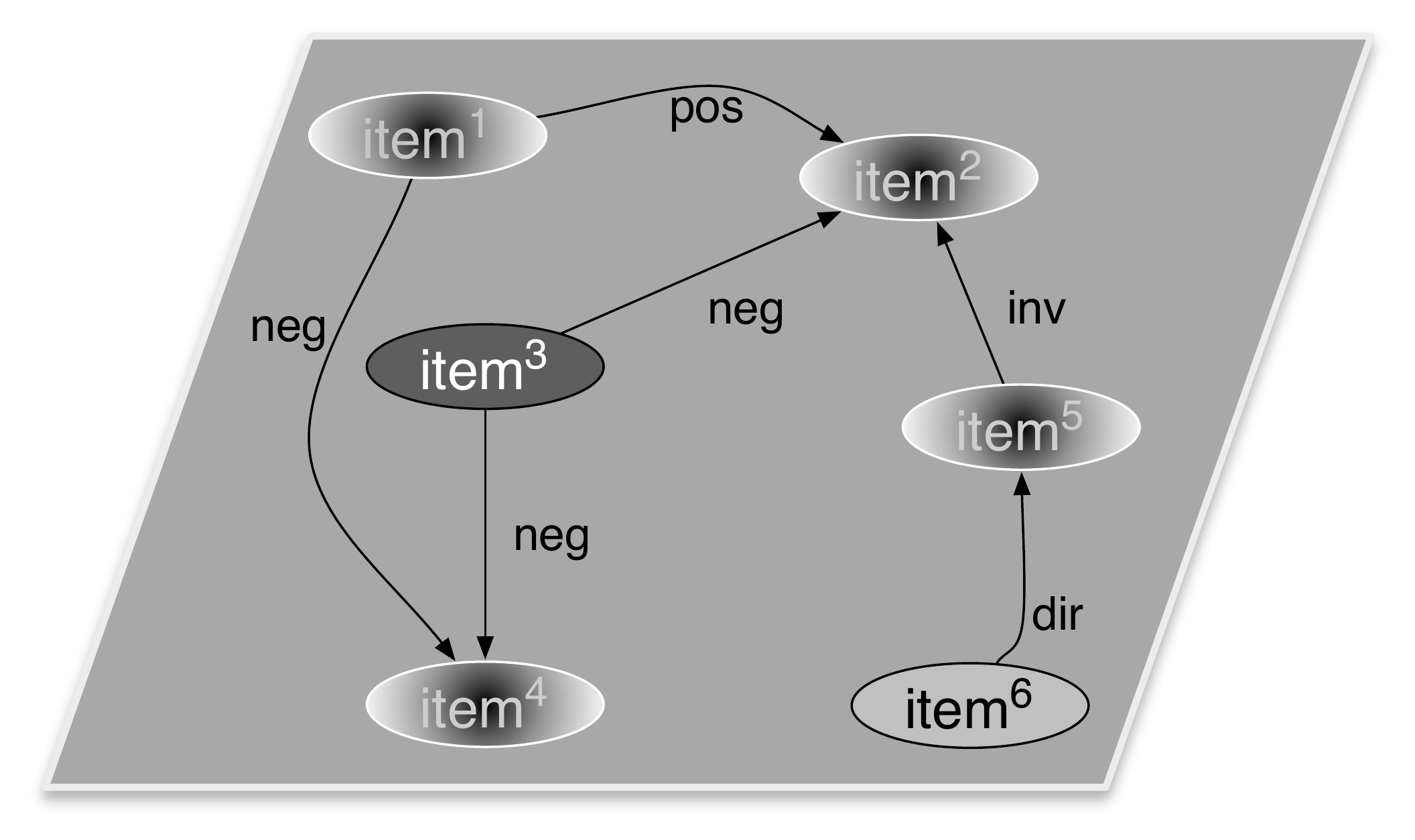}
&
\includegraphics[width=0.50\textwidth,keepaspectratio]{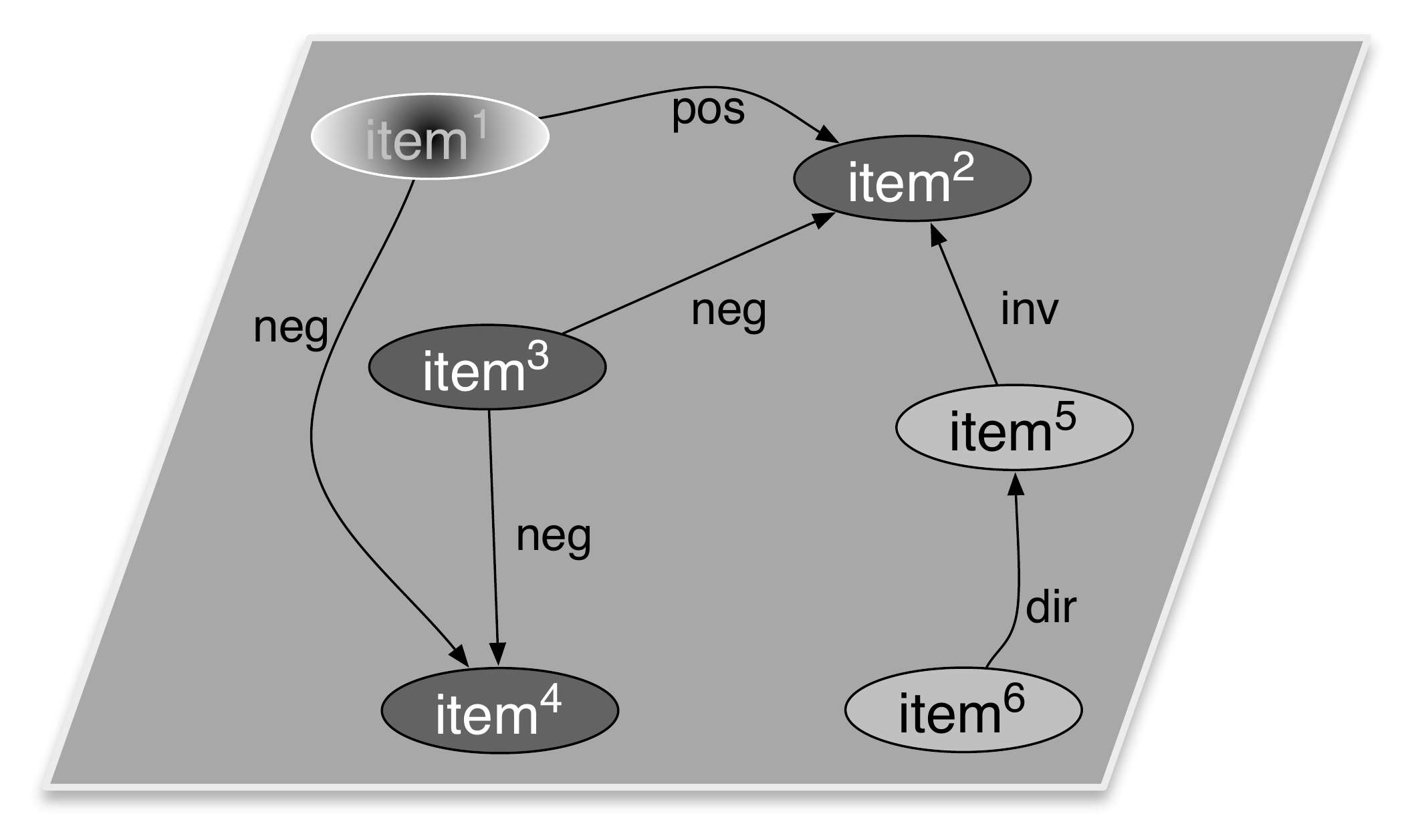}\\
\emph{Results of evaluation}
& 
\emph{Results of prediction}
\end{tabular}
\begin{tabular}{c}
\hline
\includegraphics[width=0.70\textwidth,keepaspectratio]{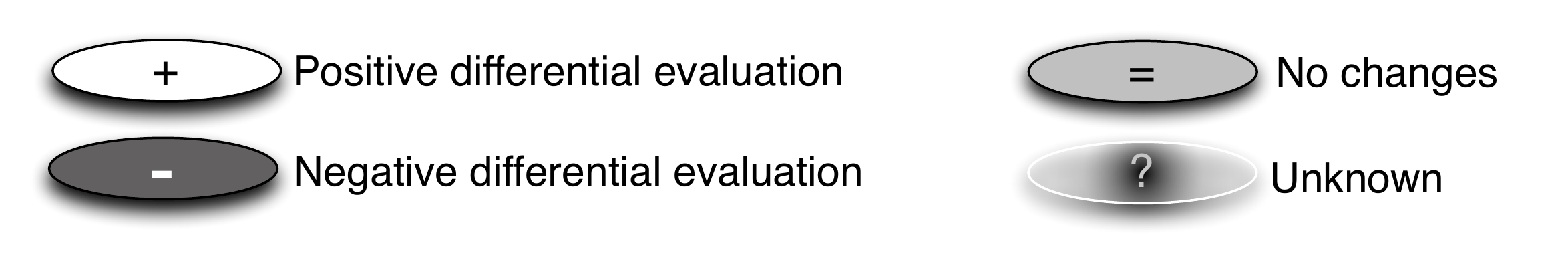}
\end{tabular}
\end{center}
\label{tab:ex-pred}
\end{table}

Since the sign of $item_1$ has three possible guesses, it is not considered as a result of prediction. All other guesses are coherent in all answer sets, thus result of the prediction task are as illustrated in the second graph in Table~\ref{tab:ex-pred}.

\end{example}

In the future, we will investigate additional optimization techniques in order to provide a more accurate qualitative evaluation of all possible consistent labeling.
Even though the following reasoning tasks consider only one solution of the prediction task, all solutions are stored for off-line evaluation. 
In the actual implementation, if there is no total consistent labeling, the solutions that allow to mark a greater number of nodes are taken into account. The way SINDI's knowledge representation model has been conceived suggests us that we can investigate inconsistent subgraphs in order to determine whether they are generated by an incomplete model (e.g. missing dependencies) or incorrect representation (imprecise or wrong dependency arc).
This is material for further study.

\item[ ]
\item[Local Explanation]

Once an optimal graph labeling has been identified as a result of prediction, the system analyses each class of items in order to identify chains of dependencies that may justify a set of predictions on items of that class.

At the moment, this reasoning process is performed on each class of items separately. This makes it possible to find the minimal common explanation to justify the results of prediction for all items of a class, but the subset of items for which we want to find minimal common explanation can be customized.

The reasoning process behaves as follows:
\begin{itemize}
  \item no matter which class of items is being evaluated, items $I_l$ that have been labelled as a result of prediction are considered as input for local explanation;
  \item starting from each item $I_l$ under consideration, if $I_l$ was assigned a label and $I_l$ is not a leaf\footnote{A node with no incoming arcs is considered a leaf.} of the graph, at least one of the incoming arcs contributing to the choice of the sign for $I_l$ should be included in the explanation;
  \item arcs are added backwards to an explanation path if the target node is reachable in the path; 
  \item an arc that justifies the attribution of a sign is not included in the explanation path for an item $I_l$ if it leads to a cycle, i.e. it has as a source a node that is already reachable in the explanation path for $I_l$;
  \item at most one explanation path for each item $I_l$ should be included in each solution;
  \item for each branch in the explanation path for an item $I_j$, if one arc $arc_1$ has already been included in the explanation path for another item $I_j'$ of the same class under consideration, it is preferred to every other arc $arc_k$ of the branch\footnote{This aspect contributes to the identification of minimal common explanations for all items of the class under consideration}.
\end{itemize}

The following example shows the results of the explanation task on a graph that has been partially labeled as a result of prediction.

\begin{example}\label{ex:explain}

Suppose we want to provide explanation for the class of items $C=\{item_3, item_4\}$, both of then labeled with a negative differential evaluation.
In the first case, the reasoning process returns two different explanation paths for $item_2$ and one path for $item_3$, represented by the dotted arcs.
If we add a dependency arc connecting $item_4$ and $item_3$ as in the graph at the bottom left of Table~\ref{tab:ex-explain}, the result is an explanation path that is partially in common for both items in $C$, since this is the preferred solution.

\begin{table}[h]
\caption{Graph for Example~\ref{ex:explain}}
\begin{center}
\begin{tabular}{cc}
\\
\includegraphics[width=0.50\textwidth,keepaspectratio]{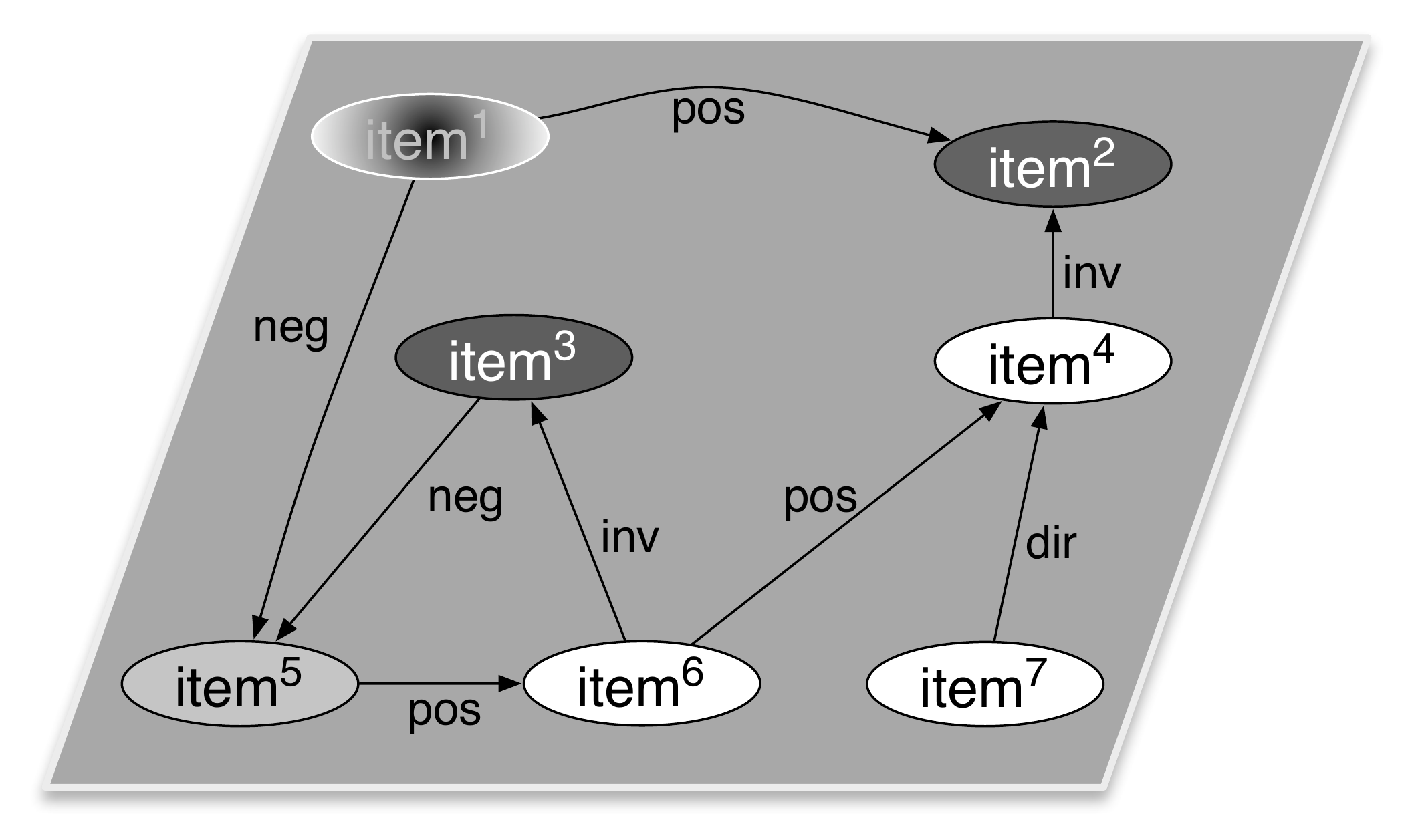}
&
\includegraphics[width=0.50\textwidth,keepaspectratio]{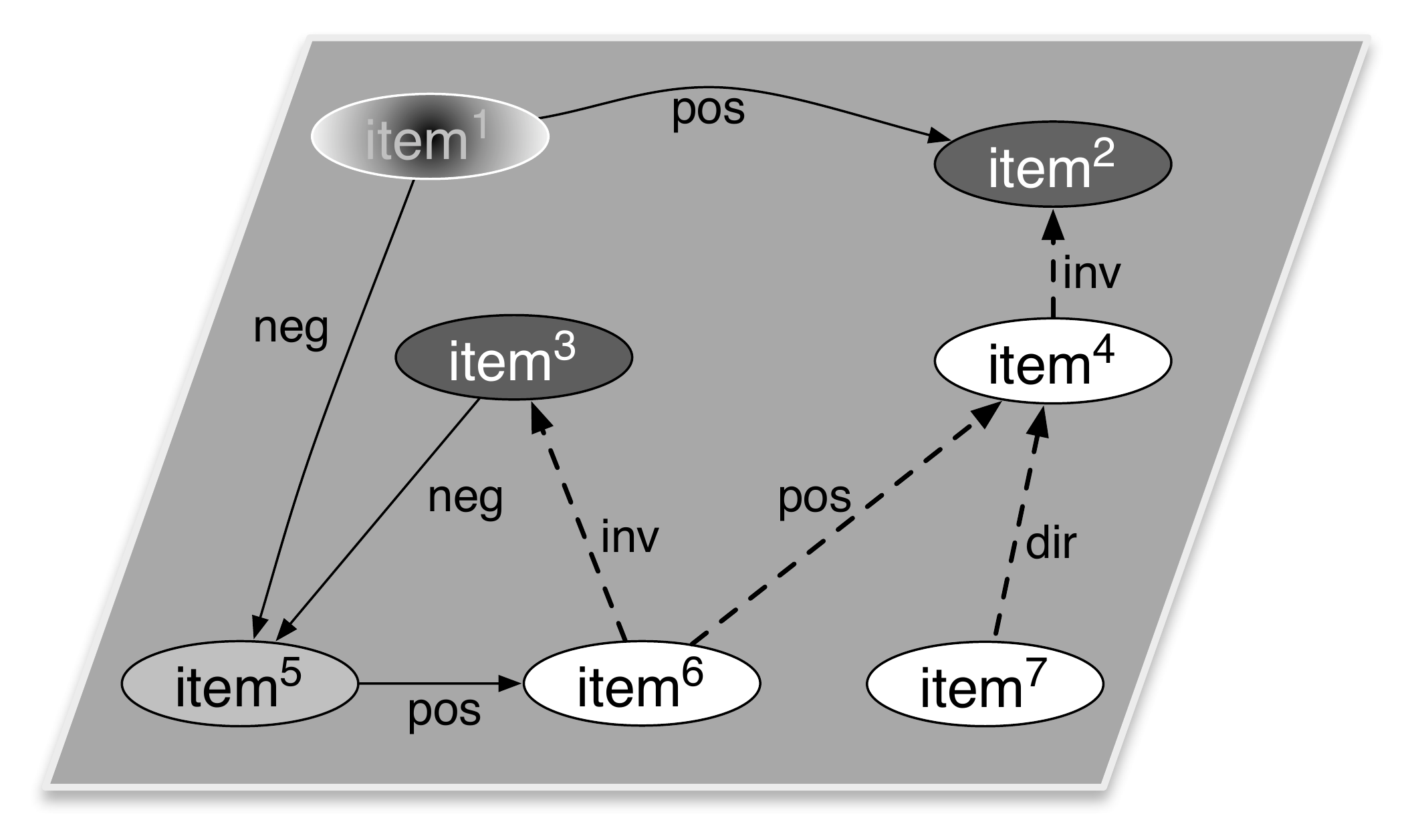}\\
\emph{Results of prediction (1)}
& 
\emph{Results of explanation (1)}\\
\hline
\includegraphics[width=0.50\textwidth,keepaspectratio]{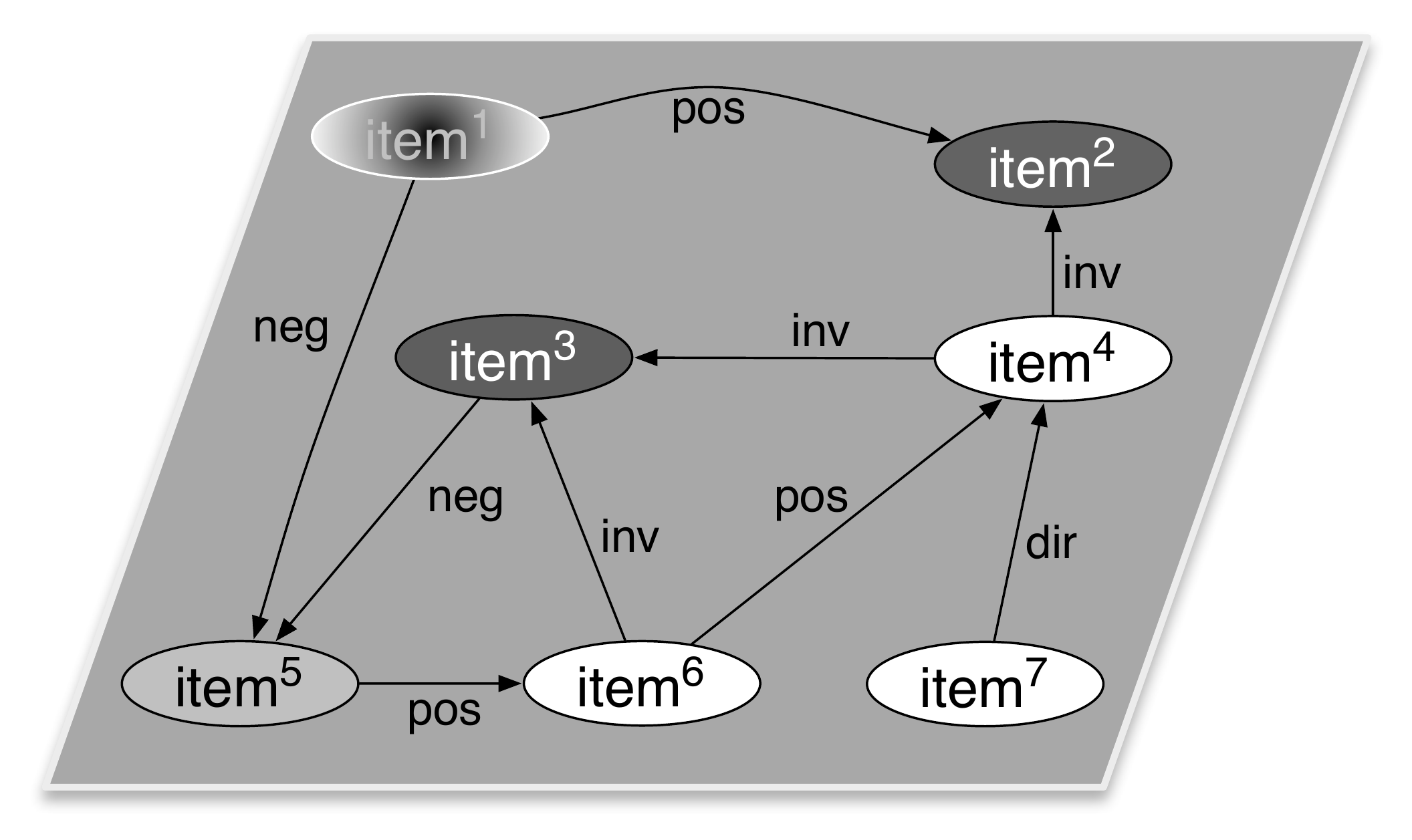}
&
\includegraphics[width=0.50\textwidth,keepaspectratio]{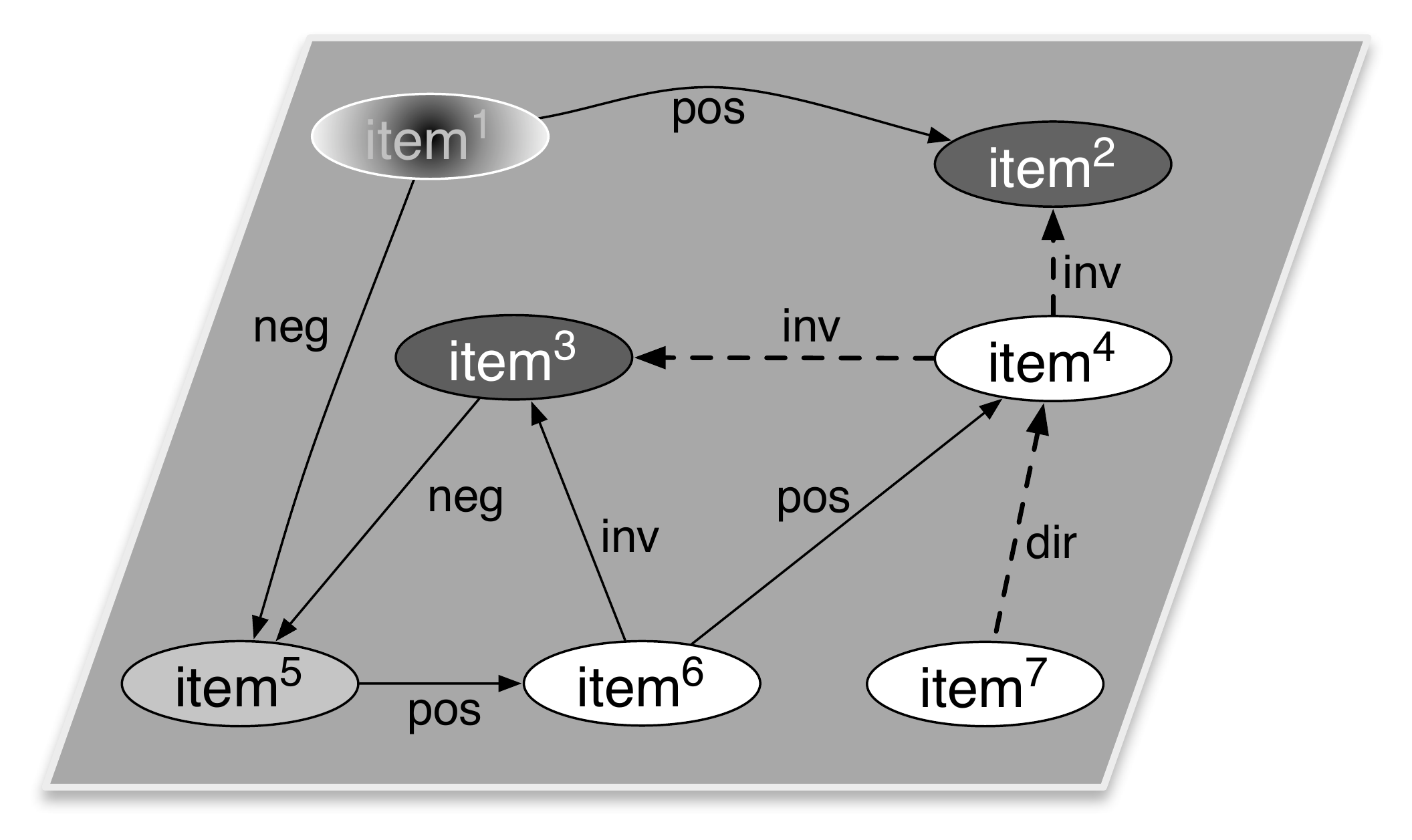}\\
\emph{Results of prediction (2)}
& 
\emph{Results of explanation (2)}
\end{tabular}
\begin{tabular}{c}
\hline
\includegraphics[width=0.70\textwidth,keepaspectratio]{./ex-legenda.pdf}
\end{tabular}
\end{center}
\label{tab:ex-explain}
\end{table}

\end{example}

\item[ ]
\item[Prevention]


In order to identify which feedback to provide, in which form and when, the reasoning component applies a set of rules we refer to as a \emph{feedback policy}.
A preliminary proposal for the application of a declarative policy description language to reason about feedback has been presented in~\cite{AC-workshop09}.
The basic idea is that a small set of well defined policy rules can be used to produce reactions of the system according to results of prediction. Since we refer to prevention as to those interventions that may keep health changes within safe boundaries, we concentrate on feedback provided as a result of predictions that identify worsening conditions.

The declarative policy language we propose admits different kinds of high-level rules that are then mapped into ASP.
These rules make it possible to represent:

\begin{itemize}
\item results of prediction for a given item $I$ identify a candidate form $f_i(X)$ for a feedback $X$ according to the explanation path associated to the prediction (event-triggering rules);
\item only one feedback form $f_i(X)$ among the triggered ones should be provided for the same feedback $X$ (choice rules);
\item a feedback form $f_i(X)$ can be preferred to form $f_k(X)$ for a given feedback $X$ under certain conditions, unless exceptions are made explicit (default ordering and exception rules);
\item the observation of an action $A_p$ performed by the person, or the selection of feedback output $f_i(X)$ triggers an action $A_s$ to be performed by system at a given time (event-condition-action rules):
\item conflicting actions cannot be executed together and qualitative optimization techniques are used to select which action to perform; as an example, we may want to select the most urgent action or the one that is better perceived by the person according to her clinical profile, or the one that minimizes interruptions (consistency rules).
\end{itemize}

Feedback outputs are used by SINDI to guide the person towards safe-behavior ans safe-living, thus helping in the application of prevention strategies. 

Each feedback action can be associated to a list of possible reactions of the patient. 
A reaction to a feedback, when detected, is logged to be used at a later time. Exploring this history, caregivers can improve the way feedback actions are performed and identify the most effective communication patterns.

Learning interaction patterns is an interesting issue~\cite{Rudary04}, but we did not tackle it yet. 
Given that we consider different forms of feedback, we could identify an appropriate reward function in order to take into account how the combination of different feedback actions and communication patterns impact on the quality of life of the person monitored.

To conclude this section, we discuss an example of how SINDI provides feedback for prevention according to the health assessment and the context in which the person acts.

\begin{example}

In this scenario, we track a hypothetical patient monitored by SINDI, called Eve. We consider SINDI's reactions aimed at reducing the risk of falls.
When Eve wakes up, SINDI tracks her getting out of bed, movements and location around the house. The quality of movements and of the environment is evaluated every hour according to the context in which data have been collected, in order to predict changes in the values of the indicators and related items.

After a long period sitting on the sofa and watching TV in the morning, Eve walks to the kitchen to prepare some food. The evaluation of her gait indicates that something has changed, since walking speed is reduced due to the long inactivity period. Trying to get things out of a cupboard, Eve slightly injures her back. This is noticed by the system because her subsequent sitting-down movements are performed with more difficulty, and the aggregation process returns a negative differential evaluation.
 
Functional dependencies are evaluated every hour, thus SINDI identifies gait, balance and mobility as problematic after analyzing indicators like walking speed and quality of sitting action in the time interval under investigation. Given that gait and balance problems directly affect mobility and that a reduced mobility may have a negative impact on the risk of falls, SINDI predicts a possible worsening of the risk of falls. 
Balance also influences dependency in getting dressed; though we have no indicators-based evidence of it, the prediction task marks dependency in this activity as subjected to worsening. 

A set of event-triggering rules identifies \emph{potential} feedback outputs among the ones in Table~\ref{tab:feedback}: stand up slowly, use a stable chair to get dressed, don't stay inactive for too long during the day, keep stairs and walk areas clear of clutter. 
An event-condition-action rule indicates that feedback outputs in form of suggestions and notifications should be provided as a report at the end of the day. This holds unless a form of higher priority is triggered for the same feedback.

Later, the quality of the environment with respect to light is marked as decreased. According to the evaluation task, a wrong use of lights may indicate visual problems, thus a possible disability in vision is inferred and added in the explanation path for the prediction of an increase of the risk of falls.

Through the specification of an appropriate event-triggering rule, another suggestion is added to the list: do not walk across a dark area.

In the following inference cycle, context interpretation reveals that Eve walked through areas that were not properly lighted.
At this point, an event triggering rule indicates that the same feedback output should now be provided in form of an alert. Given that there is a default ordering rule giving alerts higher priority than suggestions, alert becomes the preferred form for the feedback ``do not walk across a dark area''.

While suggestions are usually included in a report, alerts are usually associated with a specific action to be performed. If there are no other conflicting actions inferred by the system, the feedback is provided as indicated by the correspondent event-condition-action rule, in a way that minimizes interruptions.

This example shows how inference results predicting potential risks are used to identify the appropriate feedback for prevention, and how policy rules make this list dynamic according to how the prediction and the context evolve. In this way, SINDI can provide the most appropriate feedback at the right time. 
In fact, if the wrong use of lights was not detected, a feedback related to the use of lights would have been a suggestion in the final report, rather than an alert.

The idea behind rules of a feedback policy is that reasoning about data in a reduced temporal interval may help identifying a list of potential feedbacks as soon as some events are detected. In the following inference cycles, the new information that may be available and user/caregiver preferences are used to modify the list of feedback actions and the way they are provided to the patient. 

In the next day, the impact of feedback and the way Eve reacts to them are monitored: in case Eve reacts to one of the feedback outputs (e.g. she becomes more active), an event triggering rule can generate a Notification that is added to the daily report. Notifications are used to give evidence of i) whether Eve has followed or not feedback outputs and ii) how much does preventive intervention impact the risk of falls. This makes it possible to keep track of the whole cycle (feedback, reaction to feedback when available, impact on possible evolutions of the health state) and it can be a source of data for tests and trials to identify the correct intervention for a more general class of patients.

\end{example}

\end{description}
\section{Preliminary Evaluation}\label{sec:eval}

It is very hard to evaluate the performance of SINDI in its entirety without extensive field deployment and analysis. We are working towards this goal but do not yet have hard results from field experiments that are long enough and diverse enough to yield scientific significance. We are also working on a sort of ``canned'' evaluation of the inference engine: we generate ad-hoc sensor data by simulating the behavior of the patient with an agent simulator (Repast) and create specific situations that test the functionality of the inference. Since this evaluation is not yet completed we offer here some other evaluation data regarding the sensor network and the ASP interpreter performance.
We did several tests on the WSN and on the inference engine. The localization algorithm recognizes the correct area 90\% of the time without further filtering techniques, and movement recognition is correct 95\% of the time.

With respect to the inference engine, we evaluated ASP programs by using \emph{Gringo} as grounder and the \emph{Clasp} solver~\cite{Geb07a} that supports constraints, choice rules and weight rules~\cite{NieSim00} and can solve complex reasoning tasks very efficiently due to the heuristics used, combining ASP expressivity with boolean constraint solving.

\begin{table}
\caption{Results of ASP prediction}
\begin{center}
\begin{tabular}{c|c|c|c|c|c}
\hline
{\bf Items} & {\bf Arcs} & {\bf Labeled} & {\bf Labeled after Prediction} & {\bf Time} & {\bf Solutions} \\ 
\hline \hline
20 & 30 & 4/20 & 9/20 & 0,010 & 180 \\
20 & 30 & 10/20 & 12/20 & 0,010 & 126 \\
50 & 100 & 5/50 & 50/50 & 0,020 & 24 \\
50 & 200 & 10/50 & 48/50 & 0,030 & 3 \\
50 & 250 & 20/50 & 45/50 & 0,030 & 32 \\
50 & 250 & 35/50 & 45/50 & 0,030 & 32 \\
50 & 300 & 20/50 & 49/50 & 0,040 & 2 \\
50 & 400 & 20/50 & 50/50 & 0,040 & 1 \\
60 & 480 & 20/60 & 47/60 & 0,060 & 6 \\
60 & 480 & 30/60 & 47/60 & 0,030 & 6 \\

\hline \hline
\end{tabular}
\end{center}
\label{tab:ASPtest}
\end{table}

While the reasoning processes for context interpretation and evaluation takes advantage of the nonmonotonic nature of ASP to reason about collected and aggregated data, reasoning tasks for health assessment also rely on the computational power of ASP: the prediction task is based on a complex reasoning process that takes into account the whole graph of dependencies to find maximal consistent labellings; local explanation can consider an arbitrary class of items to find subsets of dependency arcs that represent a minimal common explanation, obtained by using qualitative and quantitative optimization techniques; prevention is based on a compact representation of feedback policies that are automatically mapped into ASP and enforced by applying adaptive optimization strategies taking into account contextual settings.


SINDI does not have to deal with graphs that are potentially very large (thousands of nodes and dependencies). The complexity for the prediction task derives from the ratio of arcs with respect to nodes (the higher the number of inner arcs for a node $I$, the higher the number of possible influences on the value of $I$). Also the number of initial observations available may influence results: the higher the number of observations, the harder it could be to find a consistent colouring when observations seem to be incoherent. Our tests on some random instances with a few hundreds of dependencies (which is still an overestimation of the real setting) showed that the connectiveness of the graph is not a problem for the ASP computation.

As for the prediction task, we summarize some results in Table~\ref{tab:ASPtest}, where time is expressed in seconds, to illustrate how reasoning can help predicting health evolutions even when few evaluations are available.
The table shows that the higher the number of labeled items, the lower is the percentage of the contribution of reasoning in labeling new items. This sometimes depends on the initial labeling, e.g. how well distributed the observed items are and how much coherent are their evaluations.

In evaluating indicators, delegating part of the aggregation process to the WSN nodes lowered the computational time by 60\% for instances of medium complexity (i.e. for a person that is active from 30 to 40 per cent of the time in a day).
We believe that the integration of ASP reasoning with constraint solving techniques~\cite{Gel08} could make context interpretation from sensor data more efficient and we plan to investigate it.
This of course does not include situations in which emergencies arise, since they are detected almost immediately.

\section{Conclusions}\label{sec:concl}

The paper has described in detail the design of a system for supporting Independent Living. The system has not yet been fully evaluated in the field but it is working and has been tested in the laboratory with real data. Full validation of systems like SINDI is very hard because the quality of the system depends on properties that are very difficult to quantize, e.g. patient and caregiver satisfaction, increase of the Healthy Life Years period, correctness in predicting risky situations, and so on. We are building the hardware necessary for a few deployments in patient's homes and are seeking funding from various agencies for large-scale field-deployment and test.	

Reasoning support to home monitoring has interesting potential developments. Among them, we want to investigate how observations that may appear to be inconsistent with the model of health can help discovering missing dependencies and refine the model itself.

Application-wise, given the high variability among trials and studies addressing prediction and prevention issues, it is still difficult to extract a coherent picture of what leads to disability and to develop coherent prevention strategies. 
In this respect, our system has the potential of automatically collecting a massive  amount of data in order to evaluate context-related prediction patterns and effective communication strategies for prevention.


\begin{thebibliography}{}

\bibitem[\protect\citeauthoryear{Akyildiz, Weilian, Sankarasubramaniam, and
  Cayirci}{Akyildiz et~al\mbox{.}}{2002}]{Akyildiz2002}
{\sc Akyildiz, I.~F.}, {\sc Weilian, S.}, {\sc Sankarasubramaniam, Y.}, {\sc
  and} {\sc Cayirci, E.~E.} 2002.
\newblock A survey on sensor networks.
\newblock {\em IEEE Communications Magazine\/}~{\em 40,\/}~8, 102--114.

\bibitem[\protect\citeauthoryear{Al-Omari and Shi}{Al-Omari and
  Shi}{2007}]{SAILNet}
{\sc Al-Omari, S.} {\sc and} {\sc Shi, W.} 2007.
\newblock Toward highly-available wsns for assisted living.
\newblock In {\em Proceedings of ACM/SIGMOBILE HealthNet 2007}. ACM Press, New
  York, NY, USA, San Juan, Puerto Rico, 31--36.

\bibitem[\protect\citeauthoryear{Bo, Han-ying, and Wen}{Bo
  et~al\mbox{.}}{2008}]{BetterLEACH}
{\sc Bo, W.}, {\sc Han-ying, H.}, {\sc and} {\sc Wen, F.} 2008.
\newblock An improved leach protocol for data gathering and aggregation in
  wireless sensor networks.
\newblock {\em Computer and Electrical Engineering, International Conference
  on\/}~{\em 0}, 398--401.

\bibitem[\protect\citeauthoryear{Boger, Hoey, Poupart, Boutilier, Fernie, and
  Mihailidis}{Boger et~al\mbox{.}}{2006}]{Coach06}
{\sc Boger, J.}, {\sc Hoey, J.}, {\sc Poupart, P.}, {\sc Boutilier, C.}, {\sc
  Fernie, G.}, {\sc and} {\sc Mihailidis, A.} 2006.
\newblock A planning system based on markov decision processes to guide people
  with dementia through activities of daily living.
\newblock {\em IEEE Transactions on Information Technology in
  Biomedicine\/}~{\em 10,\/}~2, 323--333.

\bibitem[\protect\citeauthoryear{Brewka, Niemel{\"a}, and Syrj{\"a}nen}{Brewka
  et~al\mbox{.}}{2002}]{BreNieSyr02}
{\sc Brewka, G.}, {\sc Niemel{\"a}, I.}, {\sc and} {\sc Syrj{\"a}nen, T.} 2002.
\newblock Implementing ordered disjunction using answer set solvers for normal
  programs.
\newblock In {\em Logics in Artificial Intelligence - Journées Européennes sur
  la Logique en Intelligence Artificielle}. Springer, Cosenza, Italy, 444--455.

\bibitem[\protect\citeauthoryear{Connell and Wolf}{Connell and
  Wolf}{1997}]{Con97}
{\sc Connell, B.~R.} {\sc and} {\sc Wolf, S.~L.} 1997.
\newblock Environmental and behavioral circumstances associated with falls at
  home among healthy elderly individuals.
\newblock {\em Archives of Physical Medicine and Rehabilitation\/}~{\em 78},
  179--186.

\bibitem[\protect\citeauthoryear{Fleming~K.C.}{Fleming~K.C.}{1995}]{elderly95}
{\sc Fleming~K.C., Evans~J.M., W. D. C.~D.} 1995.
\newblock Practical functional assessment of elderly persons: A primary-care
  approach.
\newblock {\em Mayo Clinic Proceedings\/}~{\em 70,\/}~9, 890--910.

\bibitem[\protect\citeauthoryear{Folstein, Folstein, and McHugh}{Folstein
  et~al\mbox{.}}{1975}]{mmt75}
{\sc Folstein, M.~F.}, {\sc Folstein, S.~E.}, {\sc and} {\sc McHugh, P.~R.}
  1975.
\newblock "mini-mental state". a practical method for grading the cognitive
  state of patients for the clinician.
\newblock {\em Psychiatric Research\/}~{\em 12,\/}~3, 189--198.

\bibitem[\protect\citeauthoryear{Gebser, Kaufmann, Neumann, and Schaub}{Gebser
  et~al\mbox{.}}{2007}]{Geb07a}
{\sc Gebser, M.}, {\sc Kaufmann, B.}, {\sc Neumann, A.}, {\sc and} {\sc Schaub,
  T.} 2007.
\newblock clasp: A conflict-driven answer set solver.
\newblock In {\em Ninth International Conference on Logic Programming and
  Nonmonotonic Reasoning}. Springer-Verlag, 260--265.

\bibitem[\protect\citeauthoryear{Gebser, Schaub, and Thiele}{Gebser
  et~al\mbox{.}}{2007}]{Geb07b}
{\sc Gebser, M.}, {\sc Schaub, T.}, {\sc and} {\sc Thiele, S.} 2007.
\newblock Gringo : A new grounder for answer set programming.
\newblock In {\em LPNMR}. 266--271.

\bibitem[\protect\citeauthoryear{Gelfond and Lifschitz}{Gelfond and
  Lifschitz}{1988}]{GelLif88}
{\sc Gelfond, M.} {\sc and} {\sc Lifschitz, V.} 1988.
\newblock The stable model semantics for logic programming.
\newblock In {\em International Conference on Logic Programming}. Seattle,
  Washington, 1070--1080.

\bibitem[\protect\citeauthoryear{Guigoz, Vellas, and Garry}{Guigoz
  et~al\mbox{.}}{1994}]{mna94}
{\sc Guigoz, Y.}, {\sc Vellas, B.}, {\sc and} {\sc Garry, P.} 1994.
\newblock Mini nutritional assessment: A practical assessment tool for grading
  the nutritional state of elderly patients.
\newblock {\em Facts and Research in Gerontology\/}~{\em 2}, 15--59.

\bibitem[\protect\citeauthoryear{Haigh and Yanco}{Haigh and
  Yanco}{2002}]{haigh02-survey}
{\sc Haigh, K.~Z.} {\sc and} {\sc Yanco, H.} 2002.
\newblock Automation as caregiver: A survey of issues and technologies.
\newblock In {\em Proceedings of the AAAI-02 Workshop ``Automation as
  Caregiver''"}. AAAI, 39--53.

\bibitem[\protect\citeauthoryear{Heinzelman, Chandrakasan, and
  Balakrishnan}{Heinzelman et~al\mbox{.}}{2000}]{Leach00}
{\sc Heinzelman, W.}, {\sc Chandrakasan, A.}, {\sc and} {\sc Balakrishnan, H.}
  2000.
\newblock Energy-efficient communication protocol for wireless sensor networks.
\newblock In {\em International Conference on System Sciences}. IEEE Computer
  Society, Washington, DC, USA, 8020.

\bibitem[\protect\citeauthoryear{Katz, Downs, Cash, and Grotz}{Katz
  et~al\mbox{.}}{1970}]{Katz70}
{\sc Katz, S.}, {\sc Downs, H.}, {\sc Cash, H.}, {\sc and} {\sc Grotz, R.}
  1970.
\newblock Progress in development of the index of adl.
\newblock {\em Gerontologist\/}~{\em 10,\/}~1, 20--30.

\bibitem[\protect\citeauthoryear{Lawton}{Lawton}{1988}]{Law88}
{\sc Lawton, M.} 1988.
\newblock Scales to measure competence in everyday activities.
\newblock {\em Psychopharmacological Bulletin\/}~{\em 24,\/}~4, 609--614.

\bibitem[\protect\citeauthoryear{Leone, Pfeifer, Faber, Eiter, Gottlob, Perri,
  and Scarcello}{Leone et~al\mbox{.}}{2006}]{Dlv06}
{\sc Leone, N.}, {\sc Pfeifer, G.}, {\sc Faber, W.}, {\sc Eiter, T.}, {\sc
  Gottlob, G.}, {\sc Perri, S.}, {\sc and} {\sc Scarcello, F.} 2006.
\newblock The dlv system for knowledge representation and reasoning.
\newblock {\em ACM Trans. Comput. Log.\/}~{\em 7,\/}~3, 499--562.

\bibitem[\protect\citeauthoryear{Liao, Patterson, Fox, and Kautz}{Liao
  et~al\mbox{.}}{2004}]{Liao04}
{\sc Liao, L.}, {\sc Patterson, D.}, {\sc Fox, D.}, {\sc and} {\sc Kautz, H.}
  2004.
\newblock Behavior recognition in assisted cognition.
\newblock In {\em Proceedings The AAAI-04 Workshop on Supervisory Control of
  Learning and Adaptive Systems}.

\bibitem[\protect\citeauthoryear{Malan, Fulford-Jones, Wesh, and Moulton}{Malan
  et~al\mbox{.}}{2004}]{CodeBlue}
{\sc Malan, D.}, {\sc Fulford-Jones, T.}, {\sc Wesh, M.}, {\sc and} {\sc
  Moulton, S.} 2004.
\newblock Codeblue: An ad hoc sensor network infrastructure for emergency
  medical care.
\newblock In {\em MobySys Workshop on Applications of Mobile Embedded Systems}.
  Boston, Massachusetts, USA, 12--14.

\bibitem[\protect\citeauthoryear{Marek and Truszczynski}{Marek and
  Truszczynski}{1989}]{Marek89}
{\sc Marek, W.} {\sc and} {\sc Truszczynski, M.} 1989.
\newblock Stable semantics for logic programs and default theories.
\newblock In {\em NACLP}. 243--256.

\bibitem[\protect\citeauthoryear{Mellarkod, Gelfond, and Zhang}{Mellarkod
  et~al\mbox{.}}{2008}]{Gel08}
{\sc Mellarkod, V.~S.}, {\sc Gelfond, M.}, {\sc and} {\sc Zhang, Y.} 2008.
\newblock Integrating answer set programming and constraint logic programming.
\newblock {\em Ann. Math. Artif. Intell.\/}~{\em 53,\/}~1-4, 251--287.

\bibitem[\protect\citeauthoryear{Mileo and Bisiani}{Mileo and
  Bisiani}{2009}]{AC-workshop09}
{\sc Mileo, A.} {\sc and} {\sc Bisiani, R.} 2009.
\newblock Context-aware prediction and prevention to extend healthy life
  years:preventing falls.
\newblock In {\em Proc. of the IJCAI Workshop on Intelligent Systems for
  Assisted Cognition, \textit{to appear}}.

\bibitem[\protect\citeauthoryear{Ness, Gurney, and Ice}{Ness
  et~al\mbox{.}}{2003}]{Nes03}
{\sc Ness, K.~K.}, {\sc Gurney, J.~G.}, {\sc and} {\sc Ice, G.~H.} 2003.
\newblock Screening, education, nd associated behavioral responses to reduce
  risk for falls among poeple over age 65 years atending a communiti health
  fair.
\newblock {\em Physical Therapy\/}~{\em 83,\/}~7, 631--637.

\bibitem[\protect\citeauthoryear{Niemel{\"a}}{Niemel{\"a}}{1999}]{Nie99}
{\sc Niemel{\"a}, I.} 1999.
\newblock Logic programs with stable model semantics as a constraint
  programming paradigm.
\newblock {\em Ann. Math. Artif. Intell.\/}~{\em 25,\/}~3-4, 241--273.

\bibitem[\protect\citeauthoryear{Niemel\"{a} and Simons}{Niemel\"{a} and
  Simons}{2001}]{NieSim00}
{\sc Niemel\"{a}, I.} {\sc and} {\sc Simons, P.} 2001.
\newblock Extending the smodels system with cardinality and weight constraints.
\newblock In {\em Logic-based artificial intelligence}. Kluwer Academic
  Publishers, Norwell, MA, USA, 491--521.

\bibitem[\protect\citeauthoryear{Pollack}{Pollack}{2005}]{Pollack05}
{\sc Pollack, M.~E.} 2005.
\newblock Intelligent technology for an aging population: The use of ai to
  assist elders with cognitive impairment.
\newblock {\em AI Magazine\/}~{\em 26,\/}~2, 9--24.

\bibitem[\protect\citeauthoryear{Pollack, Brown, Colbry, McCarthy, Orosz,
  Peintner, Ramakrishnan, and Tsamardinos}{Pollack
  et~al\mbox{.}}{2003}]{Autominder03}
{\sc Pollack, M.~E.}, {\sc Brown, L.~E.}, {\sc Colbry, D.}, {\sc McCarthy,
  C.~E.}, {\sc Orosz, C.}, {\sc Peintner, B.}, {\sc Ramakrishnan, S.}, {\sc
  and} {\sc Tsamardinos, I.} 2003.
\newblock Autominder: an intelligent cognitive orthotic system for people with
  memory impairment.
\newblock {\em Robotics and Autonomous Systems\/}~{\em 44,\/}~3--4, 273--282.

\bibitem[\protect\citeauthoryear{Rubenstein}{Rubenstein}{2006}]{Rub06}
{\sc Rubenstein, L.} 2006.
\newblock Falls in older people: epidemiology, risk factors and strategies for
  prevention.
\newblock {\em Age Ageing\/}~2, 37--41.

\bibitem[\protect\citeauthoryear{Rudary, Singh, and Pollack}{Rudary
  et~al\mbox{.}}{2004}]{Rudary04}
{\sc Rudary, M.}, {\sc Singh, S.}, {\sc and} {\sc Pollack, M.~E.} 2004.
\newblock Adaptive cognitive orthotics: combining reinforcement learning and
  constraint-based temporal reasoning.
\newblock In {\em Proceedings of ICML 2004}. ACM, 91--98.

\bibitem[\protect\citeauthoryear{Ryan, Pascoe, and Morse}{Ryan
  et~al\mbox{.}}{1998}]{Ryan98}
{\sc Ryan, N.~S.}, {\sc Pascoe, J.}, {\sc and} {\sc Morse, D.~R.} 1998.
\newblock Enhanced reality fieldwork: the context-aware archaeological
  assistant.
\newblock In {\em Computer Applications in Archaeology 1997}, {V.~Gaffney},
  {M.~van Leusen}, {and} {S.~Exxon}, Eds. British Archaeological Reports.
  Tempus Reparatum.

\bibitem[\protect\citeauthoryear{Savvides, Han, and Strivastava}{Savvides
  et~al\mbox{.}}{2001}]{Savvides2001}
{\sc Savvides, A.}, {\sc Han, C.-C.}, {\sc and} {\sc Strivastava, M.~B.} 2001.
\newblock Dynamic fine-grained localization in ad-hoc networks of sensors.
\newblock In {\em MobiCom '01: Proceedings of the 7th annual international
  conference on Mobile computing and networking}. ACM, New York, NY, USA,
  166--179.

\bibitem[\protect\citeauthoryear{Schilit, Adams, and Want}{Schilit
  et~al\mbox{.}}{1994}]{Schilit94}
{\sc Schilit, B.}, {\sc Adams, N.}, {\sc and} {\sc Want, R.} 1994.
\newblock Context-aware computing applications.
\newblock In {\em In Proceedings of the Workshop on Mobile Computing Systems
  and Applications}. IEEE Computer Society, 85--90.

\bibitem[\protect\citeauthoryear{Simons, Niemel\"{a}, and Soininen}{Simons
  et~al\mbox{.}}{2002}]{Nie02}
{\sc Simons, P.}, {\sc Niemel\"{a}, I.}, {\sc and} {\sc Soininen, T.} 2002.
\newblock Extending and implementing the stable model semantics.
\newblock {\em Artificial Intelligence Journal\/}~{\em 138,\/}~1-2, 181--234.

\bibitem[\protect\citeauthoryear{Stuck, Walthert, Nikolaus, christophe J.~Bula,
  Hohmann, and Beck}{Stuck et~al\mbox{.}}{1999}]{Stuck99}
{\sc Stuck, A.~E.}, {\sc Walthert, J.~M.}, {\sc Nikolaus, T.}, {\sc christophe
  J.~Bula}, {\sc Hohmann, C.}, {\sc and} {\sc Beck, J.~C.} 1999.
\newblock Risk factors for functional status decline in community-living
  elderly people: a systematic literature review.
\newblock {\em Social Science \& Medicine\/}~{\em 48,\/}~4, 445--469.

\bibitem[\protect\citeauthoryear{Tinetti, Williams, and Mayewski}{Tinetti
  et~al\mbox{.}}{2002}]{poma86}
{\sc Tinetti, M.}, {\sc Williams, T.}, {\sc and} {\sc Mayewski, P.} 2002.
\newblock Fall risk index for elderly patients based on number of chronic
  disabilities.
\newblock {\em American Journal of Medicine\/}~{\em 80}, 429--434.

\bibitem[\protect\citeauthoryear{Tonelli}{Tonelli}{2001}]{Ton01}
{\sc Tonelli, M.~R.} 2001.
\newblock The limits of evidence-based medicine.
\newblock {\em Respir Care\/}~{\em 46,\/}~12 (December), 1435--40.

\bibitem[\protect\citeauthoryear{Wood, Virone, Doan, Cao, Selavo, Wu, Fang, He,
  Lin, and Stankovic}{Wood et~al\mbox{.}}{2006}]{ALARM-NET}
{\sc Wood, A.}, {\sc Virone, G.}, {\sc Doan, T.}, {\sc Cao, Q.}, {\sc Selavo,
  L.}, {\sc Wu, Y.}, {\sc Fang, L.}, {\sc He, Z.}, {\sc Lin, S.}, {\sc and}
  {\sc Stankovic, J.} 2006.
\newblock Alarm-net: Wireless sensor networks for assisted-living and
  residential monitoring.
\newblock Tech. Rep. CS-2006-11, Dep. of Computer Science, University of
  Virginia.

\bibitem[\protect\citeauthoryear{Yamamoto, Mogi, Umegaki, Suzuki, Ando,
  Shimokata, and Iguchi}{Yamamoto et~al\mbox{.}}{2004}]{cdt04}
{\sc Yamamoto, S.}, {\sc Mogi, N.}, {\sc Umegaki, H.}, {\sc Suzuki, Y.}, {\sc
  Ando, F.}, {\sc Shimokata, H.}, {\sc and} {\sc Iguchi, A.} 2004.
\newblock The clock drawing test as a valid screening method for mild cognitive
  impairment.
\newblock {\em Dementia and Geriatric Cognitive Disordorders\/}~{\em 18},
  172--179.

\bibitem[\protect\citeauthoryear{Yesavage, Brink, Rose, Lum, Huang, Adey, and
  Leirer}{Yesavage et~al\mbox{.}}{1983}]{gds83}
{\sc Yesavage, J.}, {\sc Brink, T.}, {\sc Rose, T.}, {\sc Lum, O.}, {\sc Huang,
  V.}, {\sc Adey, M.}, {\sc and} {\sc Leirer, V.} 1982-1983.
\newblock Development and validation of a geriatric depression screening scale:
  A preliminary report.
\newblock {\em Journal of Psychiatric Research\/}~{\em 17,\/}~1, 37--49.

\end{thebibliography}

\end{document}